\documentclass{article}
\PassOptionsToPackage{numbers}{natbib}

\usepackage[final]{neurips_2025}

\usepackage[utf8]{inputenc} % allow utf-8 input
\usepackage[T1]{fontenc}    % use 8-bit T1 fonts
\usepackage{hyperref}       % hyperlinks
\usepackage{url}            % simple URL typesetting
\usepackage{booktabs}       % professional-quality tables
\usepackage{amsfonts}       % blackboard math symbols
\usepackage{nicefrac}       % compact symbols for 1/2, etc.
\usepackage{microtype}      % microtypography
\usepackage{xcolor}         % colors
\usepackage{graphicx}         % 我自己加的
\usepackage{amsmath}        % 我自己加的
\usepackage{multirow}        % 我自己加的
\usepackage{amssymb}        % 我自己加的
\usepackage{algpseudocode}
\usepackage{algorithm}
\usepackage{wrapfig}
\usepackage[hang,flushmargin]{footmisc}
\newcommand\blfootnote[1]{% 
    \begingroup 
    \renewcommand\thefootnote{} \footnote{#1}% 
    \addtocounter{footnote}{-1}% 
    \endgroup 
}

\title{ShoeFit: A New Dataset and Dual-image-stream DiT Framework for Virtual Footwear Try-On}

\author{
\small 
Yuhan Li$^{1*\dag}$, Zhiyu Jin$^{2*}$,  Yifan Tong$^{2}$, Wenxiang Shang$^{2}$, Benlei Cui$^{2}$, \\
\small \textbf{Xuanhong Chen$^{1}$, Hangcheng Zhu$^{2}$, Bingbing Ni$^{1\ddag}$}
\\[2pt] 
\small
$^{1}$Shanghai Jiao Tong University \hspace{2em}  $^{2}$Alibaba Group 
\\[2pt] \small 
\{melodious, nibingbing\}@sjtu.edu.cn \\
\small 
\url{https://colorful-liyu.github.io/ShoeFit/} \\
}

\begin{document}

\maketitle

\blfootnote{* Equal contributions. \\
\dag  Work done during an internship at Alibaba. \\
 \ddag  Corresponding author. \\ }

\begin{figure}[h!]
  \centering
  \vspace{-9mm}
  %\fbox{\rule[-.5cm]{0cm}{4cm} \rule[-.5cm]{4cm}{0cm}}
  \includegraphics[width=\linewidth]{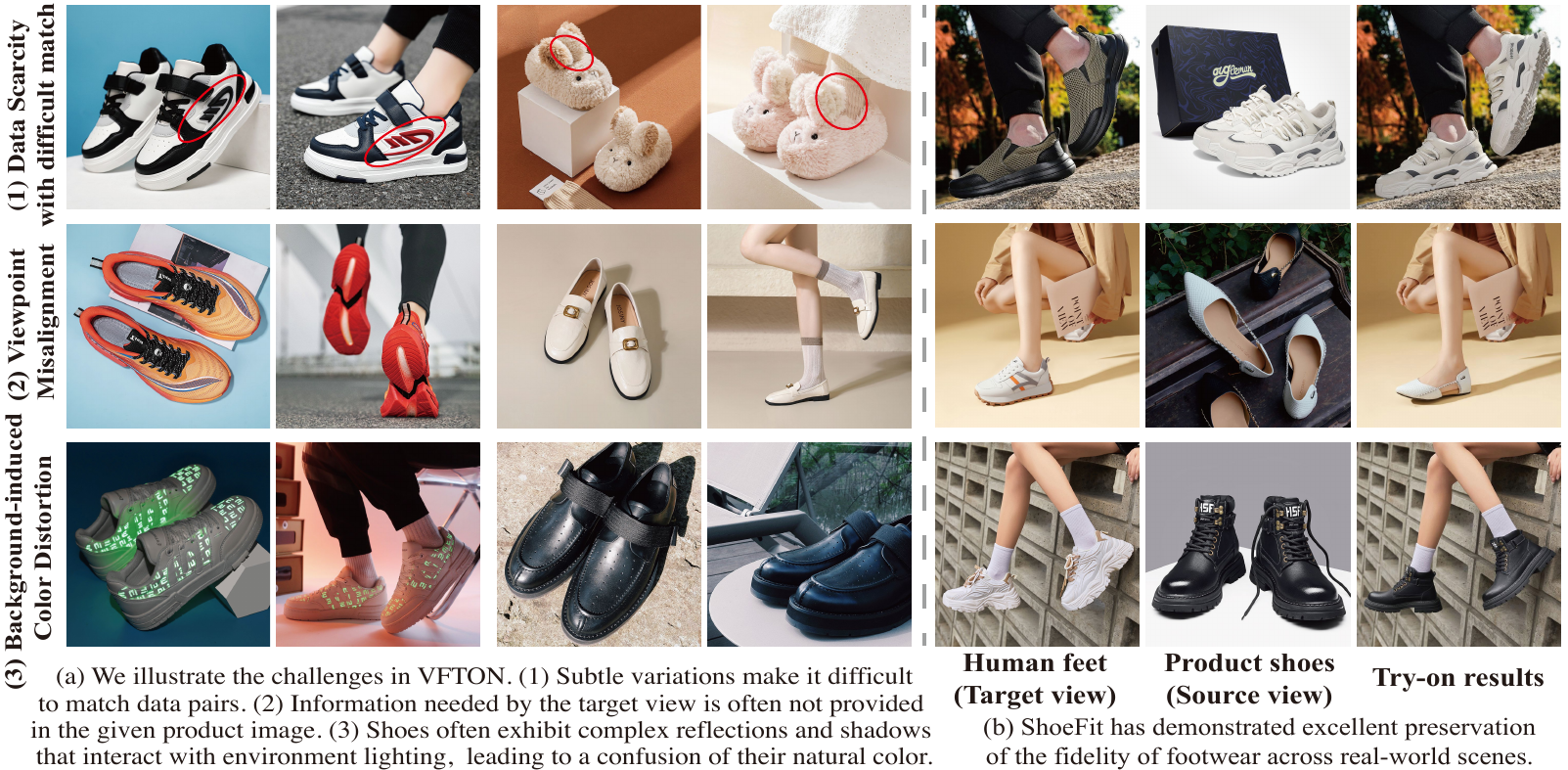}
  \vspace{-5mm}
  \caption{VFTON presents three challenges, while ShoeFit gives a successful solution.}
  \vspace{-2mm}
  \label{fig: motivation}
  
\end{figure}

\begin{abstract}
%\vspace{-3mm}
Virtual footwear try-on (VFTON), a critical yet underexplored area in virtual try-on (VTON), aims to synthesize faithful try-on results given diverse footwear and model images while maintaining 3D consistency and texture authenticity. Unlike conventional garment-focused VTON methods, VFTON presents unique challenges due to (1) \textbf{Data Scarcity}, which arises from the difficulty of perfectly matching product shoes with models wearing the identical ones, (2) \textbf{Viewpoint Misalignment}, where the target foot pose and source shoe views are always misaligned, leading to incomplete texture information and detail distortion, and (3) \textbf{Background-induced Color Distortion}, where complex material of footwear interacts with environmental lighting, causing unintended color contamination. To address these challenges, we introduce MVShoes, a multi-view shoe try-on dataset consisting of 7305 well-annotated image triplets, covering diverse footwear categories and challenging try-on scenarios. Furthermore, we propose a dual-stream DiT architecture, \textbf{ShoeFit}, designed to mitigate viewpoint misalignment through Multi-View Conditioning with 3D Rotary Position Embedding, and alleviate background-induced distortion using the LayeredRefAttention which leverages background features to modulate footwear latents. The proposed framework effectively decouples shoe appearance from environmental interferences while preserving high-quality texture detail through decoupled denoising and conditioning branches. Extensive quantitative and qualitative experiments demonstrate that our method substantially improves rendering fidelity and robustness under challenging real-world product shoes, establishing a new benchmark in high-fidelity footwear try-on synthesis.
\end{abstract}

\section{Introduction}
%%\vspace{-2mm}

Virtual Footwear Try-On (VFTON) involves synthesizing product shoes onto human foot images, preserving their geometric structure and material properties, based on input images of both the source product and the target foot. VFTON offers significant commercial potential and academic value, with Nike's annual footwear sales reaching 33,427 million dollars~\cite{nike}, while shoes, as complex 3D multi-view objects~\cite{shi2023mvdream}, present unique challenges in research on controllable generation. Unfortunately, this field has not been sufficiently explored within the community.

There are three main challenges hindering the advancement of VFTON. (1) \textbf{Data Scarcity}: Due to the prevalence of footwear with similar appearances but subtle variations, as in Fig.~\ref{fig: motivation} (a-1), such as slight differences in shoelaces, heels, and stripes, it is challenging to accurately match product shoes with models wearing the identical ones. Further exacerbating this issue, companies tend to keep data related to shoe try-ons confidential due to its clear commercial value~\cite{zhang2024mmtryon, guo2025any2anytryon}. As a result, there is currently no available open-source dataset for footwear try-ons. (2) \textbf{\emph{Viewpoint Misalignment}}: Unlike flat-lay 2D garment VTON, where both the target garment and human pose are captured from aligned front-facing perspectives, as in Fig.~\ref{fig: motivation} (a-2), virtual footwear try-on inherently involves mismatched viewpoints between in-shop product images (source view) and target human poses (target view). Specifically, information needed by the target view is often not provided in the given product image due to 3D constraints, creating an information-deficient scenario that results in detail distortion and artifacts~\cite{liu2023one, shi2023zero123++}. (3) \textbf{\emph{Background-induced Color Distortion}}: Unlike the diffuse reflection properties of knitted fabrics in clothing, which ensure visual color largely remains unaffected by the background, shoes often exhibit complex reflective properties that interact with ambient lighting~\cite{kocsis2024lightit}, making them more susceptible to visual interference from background colors, as in Fig.~\ref{fig: motivation} (a-3). This interaction between appearance and background leads to color contamination from surroundings and distorted material rendering in synthesized results.
The above issues are first identified and explicitly proposed in this paper, which make it impossible for VFTON to share the same pipeline as VTON, thus presenting VFTON with unique challenges and research value.

\begin{figure}[t]
  \centering
\includegraphics[width=1.0\textwidth]{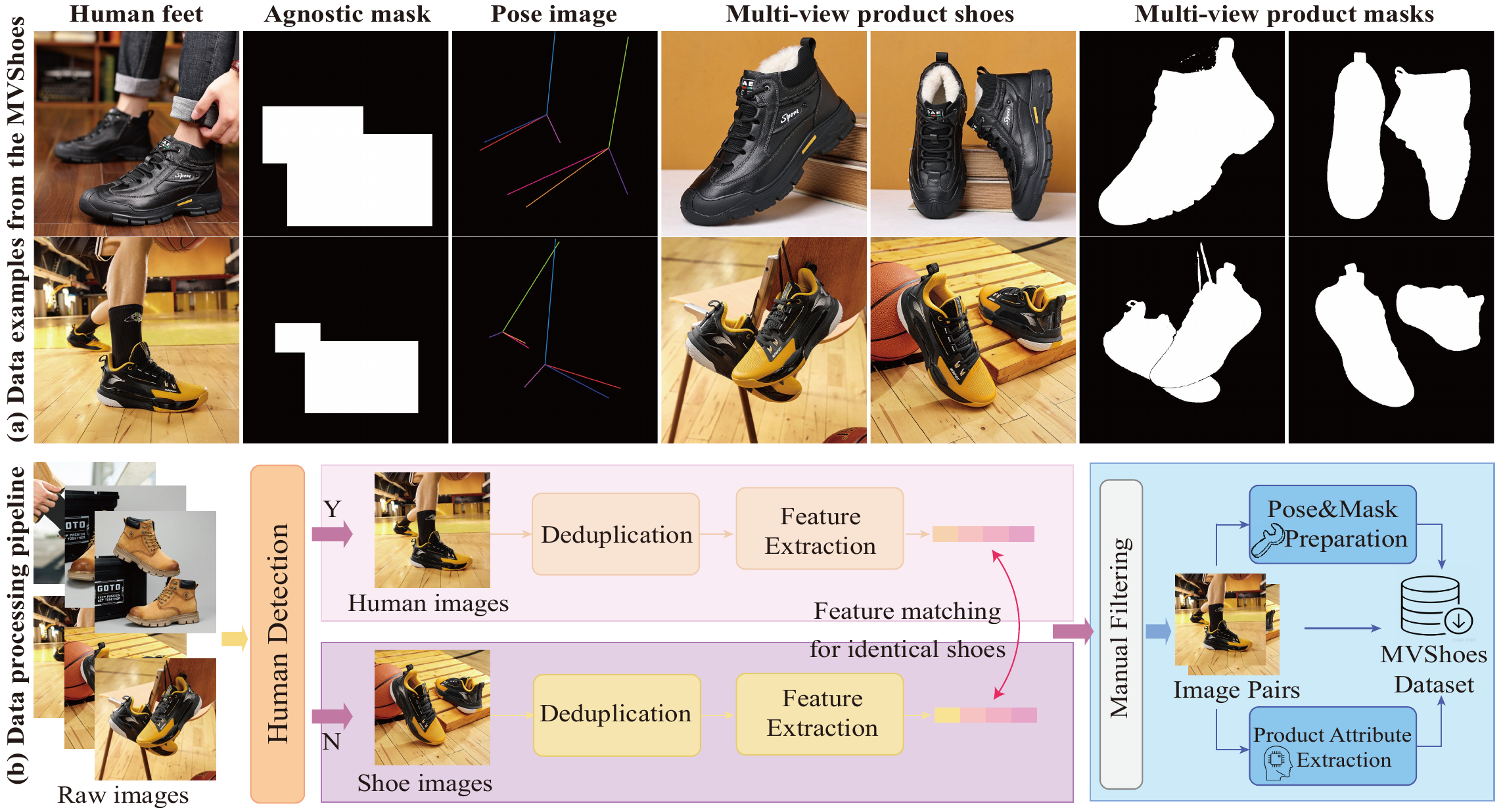}
%\vspace{-3mm}
  \caption{We illustrate examples of the MVShoes dataset in (a) and data processing pipeline in (b).} 
  %\vspace{-6mm}
  \label{fig: data_stats}
\end{figure}

To tackle these challenges, we introduce MVShoes, \textbf{the first multi-view footwear try-on dataset}. It comprises 7305 annotated high-resolution image triplets, each containing two product shoe images from different views and a corresponding try-on result, as shown in Fig.~\ref{fig: data_stats} (a). The dataset covers a diverse range of shoe categories and try-on scenarios, supported by a comprehensive and rigorous data cleaning pipeline to ensure accurate matching of identical product shoes with their human images.

Furthermore, we propose ShoeFit, a customized dual-image-stream DiT~\cite{dit} architecture for VFTON, which generates high-fidelity try-on results through decoupled denoising and conditioning branches. ShoeFit is also equipped with two key improvements:
(1) \textbf{Multi-View Conditioning}, incorporating reference concatenation~\cite{chong2024catvton} and corresponding 3D Rotary Position Embedding (3D RoPE). While a single source product view often fails to provide complete shoe texture features required for the target try-on view, the try-on results can significantly benefit from multiple source views, effectively reducing uncertainty in target surface rendering and minimizing texture distortion.
(2)\textbf{LayeredRefAttention Module}. LayeredRefAttention leverages background features to modulate shoe features~\cite{karras2019style}, decoupling the intrinsic shoe appearance from the surroundings. Additionally, the subsequent denoising attention suppresses background-related computations, effectively mitigating background color contamination. This layer-aware attention mechanism significantly enhances robustness in real-world footwear try-on scenarios.
Our contributions are summarized as follows:
%\vspace{-3mm}

\begin{itemize}
\item We formally define the VFTON problem, reporting the unique issues of Data Scarcity, Viewpoint Misalignment and Background-induced color distortion across real-world scenarios.
\item We curate the first shoe try-on dataset, containing 7305 pairs of high-quality multi-view garment-model samples, paving the way for subsequent footwear try-on studies.
\item We introduce a dual-stream architecture with Multi-View Conditioning and LayeredRefAttention modules to address the challenges in VFTON.
\item Both quantitative and qualitative experiments validate the superiority of our approach.
\end{itemize}

%\vspace{-3mm}
\section{Related Works}

\noindent \textbf{Image-based Virtual Try-on.} 
The virtual try-on task is concerned with synthesizing images of a person donning the designated garment with appropriate fit~\cite{han2018viton}, while retaining salient characteristics of the original garment and person, given a pair of images depicting a person and a target garment. To execute this task, numerous works~\cite{dong2020fashion, jo2019scfegan, lee2020maskgan, bai2022sdafn, chopra2021zflow, han2019clothflow} have utilized Generative Adversarial Networks (GANs)~\cite{goodfellow2014GAN} with two-stage strategy~\cite{lee2022hrviton, Ge2021PFAFN, yang2020acgpn}: (1) warping the clothing to the desired shape~\cite{bookstein1989tps, lin2017fpn} and (2) fusing the deformed clothing via try-on generator based on GAN. As significant progress in Text-to-Image diffusion models~\cite{ho2020ddpm, nichol2021glide, huang2023composer, chen2023anydoor} is witnessed in recent years, some works~\cite{chen2024wearanyway, gou2023dci, morelli2023ladi, kim2023stableviton, li2024anyfit} have been motivated to incorporate pre-trained diffusion models~\cite{rombach2022ldm,podell2023sdxl} as priors into virtual try-on task.
Most recently, OOTDiffusion~\cite{xu2024ootdiffusion}, IDM~\cite{choi2024idm}, and MMTryon~\cite{zhang2024mmtryon} achieve garment feature extraction with a parallel U-Net and feed them through self-attention for enhanced integration. CatVTON~\cite{chong2024catvton} concatenates the garment and person images along the spatial dimension and feed them into a single UNet. 
Unfortunately, these methods serve for fabric clothing, intrinsically lack support for faithful try-on to 3D multi-view shoes with various materials, exhibiting severe color contamination from the background.

\noindent \textbf{Controllable Diffusion Models.} 
To attain conditional control in diffusion models, T2I-Adapter~\cite{mou2024t2iadapter} and IP-Adapter~\cite{ye2023ipadapter} incorporate additional trainable modules to fuse condition features. Additionally, some investigations utilize various prompt engineering techniques~\cite{li2023gligen, yang2023reco, zhang2023controllablegpt4} and implement cross-attention constraints~\cite{chen2024layoutcontrol, xie2023boxdiff, ju2023humansd, zhang2024boow} to facilitate more controllable generation. Recently, the release of FLUX~\cite{flux} and SD3~\cite{sd3} has sparked renewed interest in DiT architectures~\cite{dit} with subsequent conditional frameworks such as Flux.1 Fill~\cite{flux-fill} and Flux.1 Redux~\cite{flux-redux} achieving inpainting and object injection capabilities through enhanced encoder designs. However, these methods remain constrained by coarse-grained conditioning mechanisms, which fail to ensure generation fidelity—particularly in VFTON tasks with high geometric variability and stringent fidelity requirements.

\section{MVShoes Dataset}
\label{sec: dataset}
We collect the first footwear try-on dataset which contains a total of 7305 annotated high-resolution image triplets, each consisting of two product shoe images from different views, and a corresponding try-on result. The dataset features diverse shoe categories and includes foot-specific pose landmarks and shoes-agnostic masks to facilitate precise alignment.

\subsection{MVShoes Collection}
All images are sourced from the Internet and include a diverse range of model scenarios such as full-body models, half-body models, foot close-ups, as well as a comprehensive variety of shoe categories, effectively meeting practical application needs. Initially, we collected approximately 25,000 raw shoe product image sets, each consisting of 2 to 5 images per shoe, including standalone product images and images of people wearing the shoes from different perspectives.

\textbf{Data Processing.}
We first employ Qwen-VL~\cite{bai2023qwenvlversatilevisionlanguagemodel} to differentiate between product footwear images (without humans) and try-on images featuring human models. Subsequently, we use the CLIP model~\cite{radford2021clip} to extract image features and set a similarity threshold to eliminate duplicate images within the same raw image set. Next, we segment the shoe region in each image, match the DINO~\cite{liu2024groundingdinomarryingdino} features and assess their similarity to filter those data pairs with different footwear colors and styles. This process results in the formation of preliminary human-footwear try-on triplets, followed by manual filtering to eliminate any errors or oversights.
We then remove visually blurred images and compile statistics on shoe categories and try-on scenarios. Finally, following the method proposed in DWPose~\cite{yang2023effectivewholebodyposeestimation} and Grounding DINO~\cite{liu2024groundingdinomarryingdino}, we extract foot poses and shoe-agnostic masks. This process culminates in the creation of high-resolution, category-comprehensive try-on triplets. The data processing pipeline is illustrated in Fig.~\ref{fig: data_stats} (b).

\begin{wrapfigure}{r}{0.5\textwidth}
%\vspace{-15mm}
  \centering
  \includegraphics[width=0.48\textwidth]{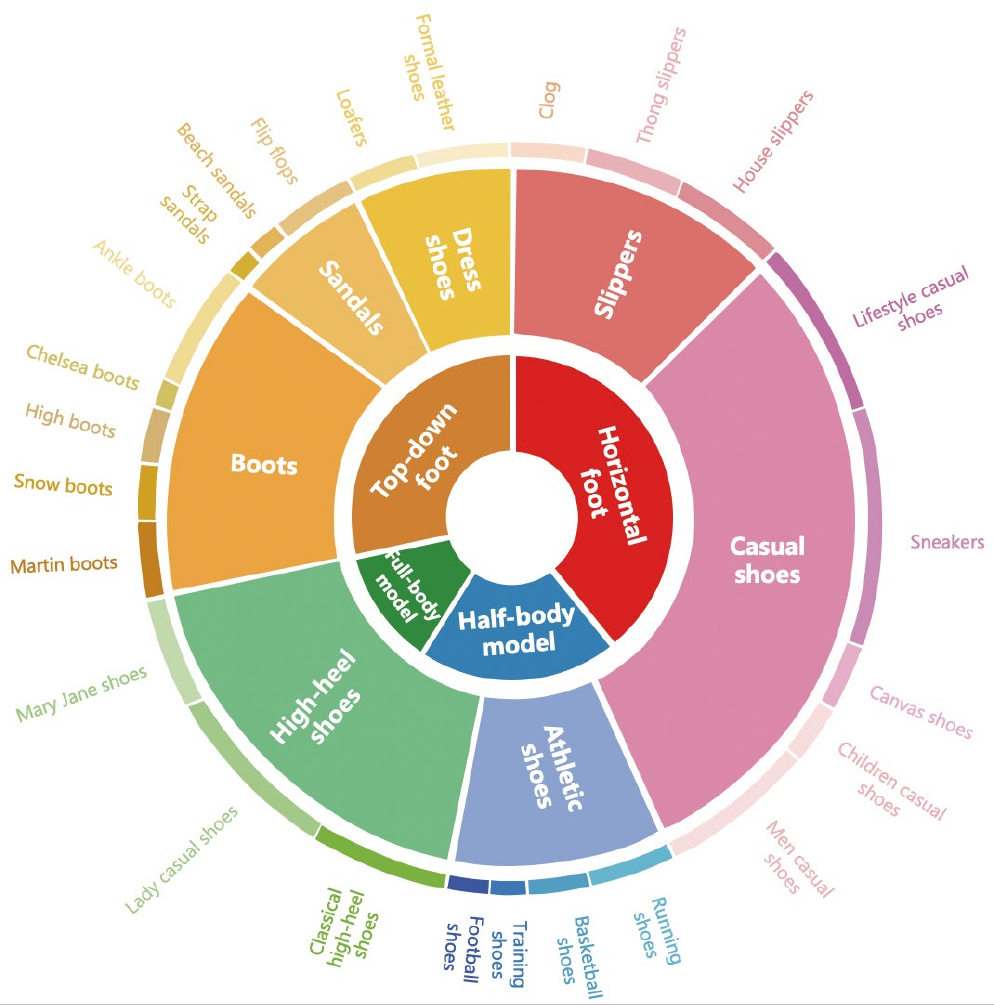}
  %\vspace{-2mm}
  \caption{The category distribution of MVShoes.} 
  \label{fig: circle}
  %\vspace{-6mm}
\end{wrapfigure}

\subsection{Dataset Statistic}

\textbf{Rich Scenes.} In conventional VTON tasks~\cite{lee2022hrviton, morelli2022dresscode}, model images typically consist of full-body or upper-body views. However, given that footwear occupies a relatively small proportion of the body in VFTON tasks, full-body and half-body models are less effective in showcasing shoe details. To address this, we simulate the distribution of footwear model scenes observed on e-commerce platforms and curate 4 model scenarios: top-down foot close-up, horizontal foot close-up, half-body model, and full-body model, as illustrated in Fig.~\ref{fig: circle}, where horizontal foot close-ups constitute the largest proportion. In such cases, accurate preservation of side details is particularly challenging, especially when compounded by viewpoint misalignment.

\textbf{Diverse Footwear Categories.} We also prioritize the comprehensiveness of footwear categories. By annotating shoe attributes~\cite{achiam2023gpt}, we establish a two-level product category label set and ensure that its distribution aligns with everyday use, as depicted in Fig.~\ref{fig: circle}. Notably, we observe that environmental lighting often affects the visual appearance of footwear, where reflections introduce background color contamination, particularly in categories such as boots, leather shoes, and high heels.

\section{Method}

\begin{figure}[t]
  \centering
\includegraphics[width=1.0\textwidth]{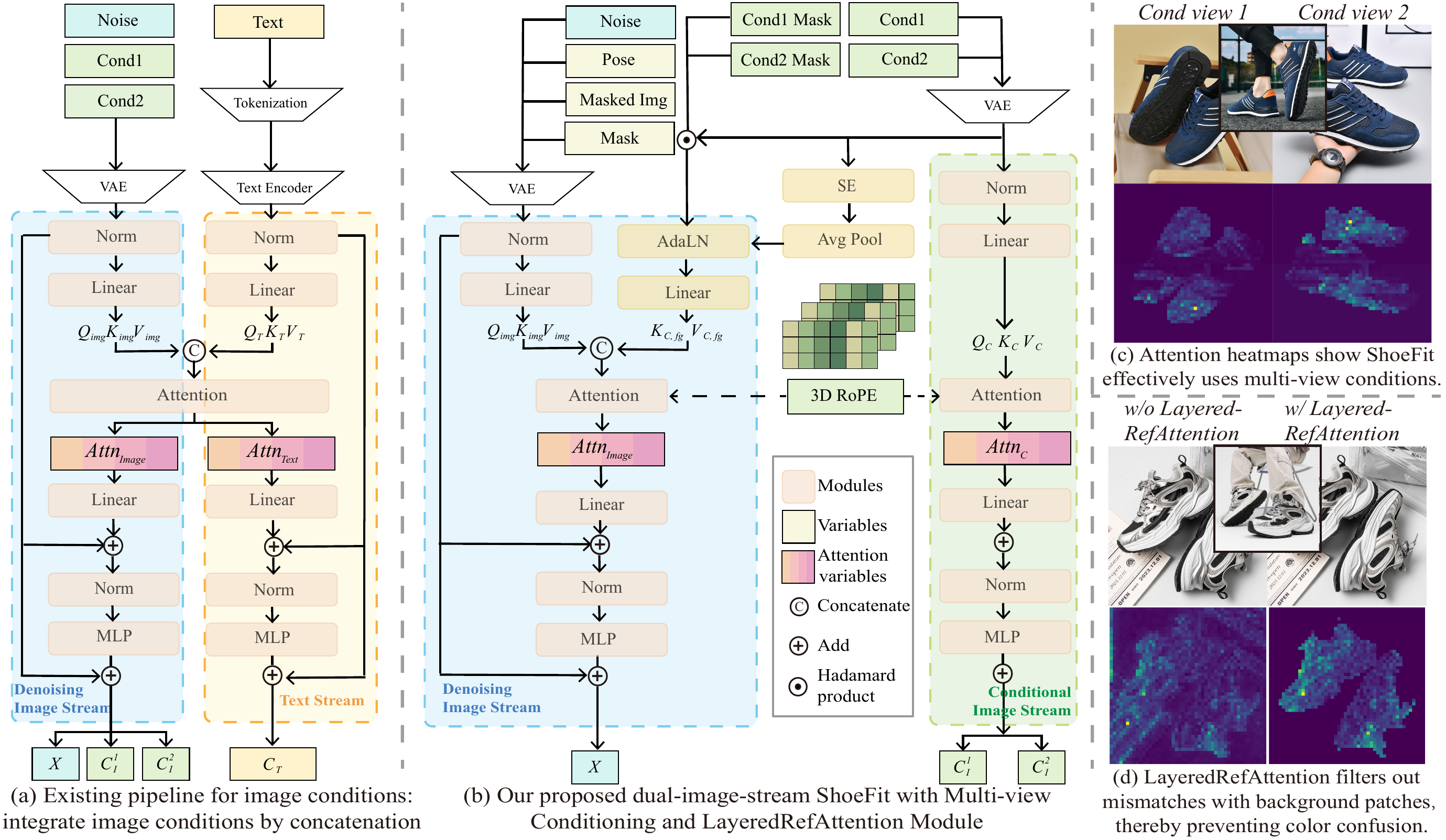}
%\vspace{-3mm}
  \caption{We illustrate existing and our DiT framework comparison in (a)-(b), and heatmaps visualization for component validation of Multi-view Conditioning and LayeredRefAttention in (c)-(d).} 
%\vspace{-4mm}
  \label{fig: framework}
\end{figure}

An overview of the ShoeFit is presented in Fig.~\ref{fig: framework} (b). Given a human image $X$ and target shoe images $C_I^{(n)}$, ShoeFit is aimed to generate an authentic try-on image. The backbone of ShoeFit employs the FLUX.1-Lite~\cite{flux}, with a customized dual-image-stream architecture in Sec.~\ref{sec: dual-stream}: denoising image stream and a copy of it as as conditioning branch to replace text modality. The features of human images and multi-view shoes are then integrated through Reference Concatenation and LayeredRefAttention modules, which are described in Sec.~\ref{sec: multi-view} and ~\ref{sec: layered-attention}.

\subsection{Dual-image-stream DiT Framework}
\label{sec: dual-stream}

The original FLUX~\cite{flux} is a text-to-image model composed of a series of stacked MM-DiT blocks. Earlier works~\cite{tan2024ominicontrol, wang2025unicombine}  directly concatenate Image/Text condition tokens with noisy image tokens such as $[X; C_{I}; C_{T}]$, relying on a single denoising stream to finish condition extraction and denoising, while the other deal with text modality, as shown in Fig.~\ref{fig: framework}(a). However, for VFTON, the generated try-ons are primarily determined by the given shoe images rather than limited semantic prompts. Thus, we remove the text stream from FLUX, achieving approximately 40\% parameter savings, including those for T5~\cite{raffel2023exploringlimitstransferlearning} and CLIP models~\cite{radford2021clip}. Instead, we replicate a portion of the denoising stream’s architecture and weights to initialize the conditioning image stream, establishing a dual-image-stream framework. This design decouples feature preservation of shoe details from the high-quality generation process, allocating additional effective model capacity to enhance overall fidelity performance compared to direct token concatenation approaches~\cite{tan2024ominicontrol,wang2025unicombine}.

\subsection{Multi-view Conditioning}
\label{sec: multi-view}

In VFTON, the information about shoes from the try-on view remains incomplete due to view mismatch and 3D-to-2D projection occlusion~\cite{3D_projection}. The inherent misalignment between source product images and target human images leads to unfaithful and physically implausible try-on results. Observing that e-commerce platforms often provide multiple-view product images (\textit{e.g.}, front, side, top views), we introduce a multi-view shoe image dataset and propose a multi-view conditional injection mechanism via Reference Concatenation and 3D RoPE. The additional perspectives comprehensively encompass more aspects of the shoe, as illustrated in Fig.~\ref{fig: framework} (c), significantly reducing uncertainty during generation and delivering high-fidelity and commercially viable synthetic results.

\textbf{Reference Concatenation.}
Inspired by CatVTON~\cite{chong2024catvton}, we adopt a simple yet effective multi-view injection strategy: concatenate multi-view conditioning shoe images along the spatial dimension as: $C_I^{cat} = [C_I^1, C_I^2, ...C_I^N]\in \mathbb{R}^{b\times N\cdot l\times c}$, where $b$ s the batch size, $l$ denotes the number of image tokens per conditioning image, and $c$ is the feature channel. The concatenated multi-view latent $C_I^{cat}$ is then fed into the conditioning image stream for feature extraction and interaction with the target human image. Concurrently, the denoising stream receives 4 concatenated components with $16+16+1+16$ channels: the noisy human latents $X_t$, the latent shoe-agnostic masked image $E(X_{masked})$, the resized agnostic mask $X_m$, and the foot pose landmarks $E(X_{pose})$, where $E(\cdot)$ represents VAE~\cite{kingma2013vae} encoding. The effectiveness of Reference Concatenation stems from the global attention interactions in attention layers. Intuitively, during the attention, each \textit{Query} patch in the human image accesses and matches the most semantically relevant \textit{Key-Value} pairs from the full set of concatenated multi-view product features, as illustrated in Fig.~\ref{fig: framework} (c). As a result, our method ensures consistent alignment between all regions in the target pose and their corresponding details in the product images, thereby achieving high-fidelity virtual try-on.

\textbf{3D RoPE.} 
However, naive Reference Concatenation introduces position misalignment in the relative positional relation of conditioning images. FLUX employs an essentially 2D RoPE scheme~\cite{su2024roformer} to encode spatial relative relationships along the height and width dimensions, where the inner product in attention depends solely on the token embeddings $x_m, x_n$ and their relative distances $m-n$:
\begin{equation}
\left \langle f_q(x_m, m), f_k(x_n, n) \right \rangle =g(x_m, x_n, m-n).
\end{equation}
When multi-view conditioning images are naively concatenated, discontinuous semantic jumps emerge at the concatenation boundaries, where tokens with small relative distances may correspond to semantically distant regions (\textit{i.e.}, different views of the footwear).  

To address this issue, we extend the 2D RoPE to three dimensions. Specifically, $E_{posID} = \text{concat}[E_{posID}^n; E_{posID}^h; E_{posID}^w] \in \mathbb{R}^{H\times W\times 3}$, where the first channel $E_{posID}^n$ serves as a \textit{index mask} to separate different conditioning views, while the second and third channels $E_{posID}^{\{h, w\}}$ encode spatial positions within each image along the height and width dimensions, respectively. Within each channel of $E_{posID}^{\{n, h, w\}}$, frequency encoding is performed given a base frequency of $\theta=10000$. 
\begin{equation}
\omega _{d}=\frac{1}{\theta ^{2d/D}}, \text{for  } d=0, 1, ..., 
\frac{D}{2}-1, \quad v_{\left \{ n, h, w \right \} } = E_{posID}^{\left \{ n, h, w \right \} } \cdot \omega_d  \in \mathbb{R}^{H\times W\times \frac{\left \{ D_n, D_h, D_w \right \}}{2} }, 
\end{equation}
where hype-parameters $D_n, D_h, D_w$ is predefined as $16, 56, 56$. Subsequently, we obtain positional rotation matrices $R_{1D}$ for each of the $n, h, w $ channels internally. By concatenating these matrices along $D$ dimension, we construct 3D RoPE matrix $R_{3D} \in \mathbb{R}^{H\times W\times  \frac{(D_n + D_h + D_w )}{2}\times 2\times 2}$ as follow:
\begin{equation}
R_{1D}^{\left \{ n, h, w \right \} }  = 
\begin{bmatrix}
cos(v_{\left \{ n, h, w \right \} } & -sin(v_{\left \{ n, h, w \right \} }) \\ 
sin(v_{\left \{ n, h, w \right \} }) & cos(v_{\left \{ n, h, w \right \} })
\end{bmatrix}
, \quad R_{3D} = \text{concat}[R_{1D}^n; R_{1D}^h; R_{1D}^w]
\end{equation}

\subsection{LayeredRefAttention Module}
\label{sec: layered-attention}

We observe that the appearance of the synthesized shoes is highly susceptible to contamination from the background colors of the conditioning images. This phenomenon arises from the complex reflective material of footwear interacting strongly with environmental lighting, termed \textbf{Background-induced Color Distortion}. Specifically, the issue arises when \textit{Query} patches from the human image inadvertently match \textit{key-value} pairs originating from the background regions of the conditioning image, as illustrated in Fig.~\ref{fig: framework} (d). A naive solution involves masking out background regions during attention, but this would \emph{\textbf{neglect critical illumination cues embedded in the background}}, which are essential for \emph{\textbf{accurately inferring the shoe’s intrinsic appearance and texture}}~\cite{kocsis2024lightit}.

To address this challenge, we propose a layer-aware attention module called \textbf{LayeredRefAttention} to explicitly differentiate between background and foreground content in the conditioning images, as illustrated in Fig.~\ref{fig: framework} (b). By leveraging background features to modulate foreground shoe features, our method decouples the intrinsic shoe appearance from its background context~\cite{karras2019style}. Subsequent denoising stream attention computations are restricted to interactions between the foreground shoe content and the target human image, effectively eliminating background illumination effects while avoiding background color contamination. Specifically, for $n\in \{1, ..., N\}$ conditioning view $C_I^{(n)}$, we employ a Squeeze-and-Excitation (SE) block ~\cite{hu2018squeeze} followed by global average pooling over spatial dimensions to compute channel-specific weights: 
\begin{equation}
P^{(n)}=AvgPool(SE(C_I^{(n)})\in \mathbb{R}^{b\times c}.
\end{equation}
Subsequently, we employ two linear layers $F(\cdot)$ to extract background modulation parameters and modulate foreground shoe features, filtering irrelevant environmental lighting and background reflection to ensure faithful material preservation of shoes as:
\begin{equation}
\beta_{scale}^{(n)}, \beta_{shift}^{(n)} = F(P^{(n)}), \quad C_{fg}^{(n)} = (1+\beta_{scale}^{(n)})\cdot (LN(C_I^{(n)} \odot M_{fg}^{(n)})) +  \beta_{shift}^{(n)}, 
\end{equation}

where $LN(\cdot)$ means layer normalization and $M_{fg}^{(n)}$ represents the binary shoe masks for conditioning product image. The attention variables $\{ K, V\}_{c, fg}^{(n)}$ extracted from $N$ views of modulated foreground conditioning feature are concatenated with the corresponding $\{Q, K, V\}_{img}$ from human image for subsequent denosing-stream attention:
\begin{equation}
Q_{C, fg}^{(n)}, K_{C, fg}^{(n)}, V_{C, fg}^{(n)} = QKV(C_{fg}^{(n)}), 
\end{equation}
\begin{equation}
Attn_{image} = \text{softmax}(\frac{Q_{img} \cdot \text{concat}(K_{img}; K_{C, fg}^{1, ... ,N, \top })} {\sqrt{d}})\cdot \text{concat}(V_{img}; V_{C, fg}^{1, ... ,N}).
\end{equation}

Meanwhile, the conditioning-stream branch performs attention computation using unmasked conditioning latents as:
\begin{equation}
Q_{C}^{(n)}, K_{C}^{(n)}, V_{C}^{(n)} = QKV(LN(C_I^{(n)})), \quad Attn_{C} = \text{softmax}(\frac{Q_{C}^{1, ... ,N} \cdot K_{C}^{1, ... ,N, \top }} {\sqrt{d}})\cdot V_{C}^{1, ... ,N}.
\end{equation}

To clarify our contributions, we omit the distillation strength normalization and final residual connections in the attention layer of FLUX.1-Lite~\cite{flux} that remain unchanged from the original implementation. In summary, LayeredRefAttention effectively enables material-background decoupling by modulating foreground shoe features to recover intrinsic color properties, resulting in texture-preserving footwear try-on synthesis, as illustrated in Fig.~\ref{fig: framework} (d).

\section{Experiments}

\begin{figure}[t]
  \centering
\includegraphics[width=1.0\textwidth]{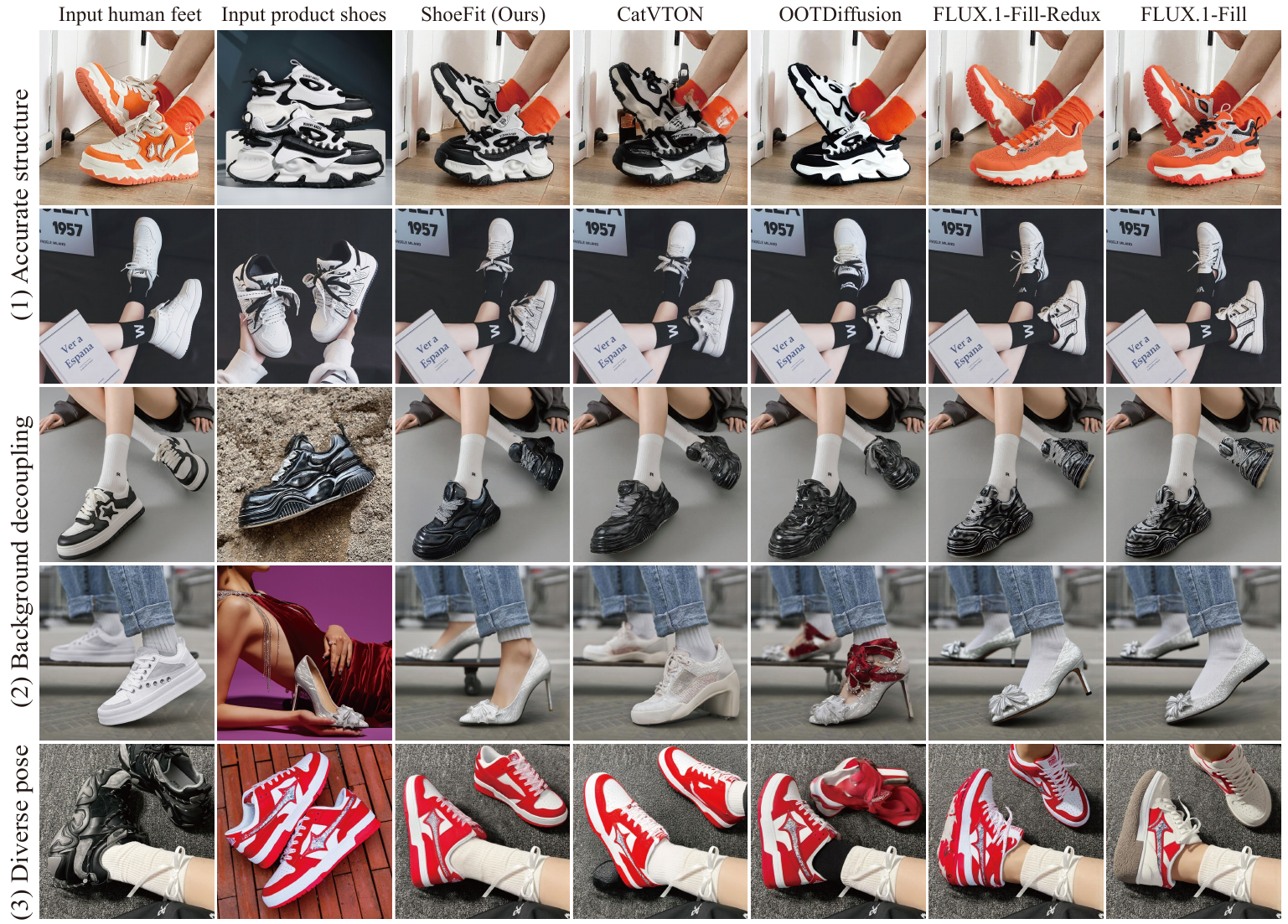}
%\vspace{-4mm}
  \caption{Comparisons on the hard scenes, where ShoeFit delivers high-fidelity and rubost results.} 

  \label{fig: main}
\end{figure}

\begin{table*}[t]
\centering
\caption{Quantitative comparison on MVshoes. ShoeFit significantly surpasses all baselines.} % 使用 caption* 使其不计入表格编号

\begin{tabular}{llccccc}
\toprule
View Num.                & Method                  &  SSIM $\uparrow$ &  LPIPS $\downarrow$  & FID $\downarrow$ & KID $\downarrow$  & DISTS $\downarrow$ \\ 
\cmidrule(lr){1-2}\cmidrule(lr){3-7}
\multirow{5}{*}{1-View} 
  & FLUX.1-Fill~\cite{flux-fill}                    & 0.725    & 0.201    & 24.44      & 2.221  & 0.1293  \\
  & FLUX.1-Fill-Redux~\cite{flux-redux}            & 0.724    & 0.199    & 24.67  & 2.086 & 0.1274  \\    
  & OOTDiffusion-SDXL~\cite{xu2024ootdiffusion}    & 0.724   & 0.194   & 22.60    & 2.345  & 0.1114  \\
  & CatVTON-Flux-Lite~\cite{chong2024catvton}    & 0.745    & 0.185    & 22.71    & 3.327  & 0.1179  \\
  & ShoeFit-1V        & \underline{0.772}    & \underline{0.159}   & \underline{21.07}    & \underline{0.944}  & 0.1059   \\
  \cmidrule(lr){1-2}\cmidrule(lr){3-7}
\multirow{2}{*}{2-View}  
  & CatVTON-Flux-Lite-2V~\cite{chong2024catvton} & 0.756 & 0.171  & 22.64 & 3.318  & \underline{0.1040}  \\ 
  & \textbf{ShoeFit-2V}  & \textbf{0.780}  & \textbf{0.149}  & \textbf{20.18} & \textbf{0.485}  & \textbf{0.0941} \\
\bottomrule
\end{tabular}

\label{tab: main}
\end{table*}

\begin{figure}[t]
  \centering
\includegraphics[width=1.0\textwidth]{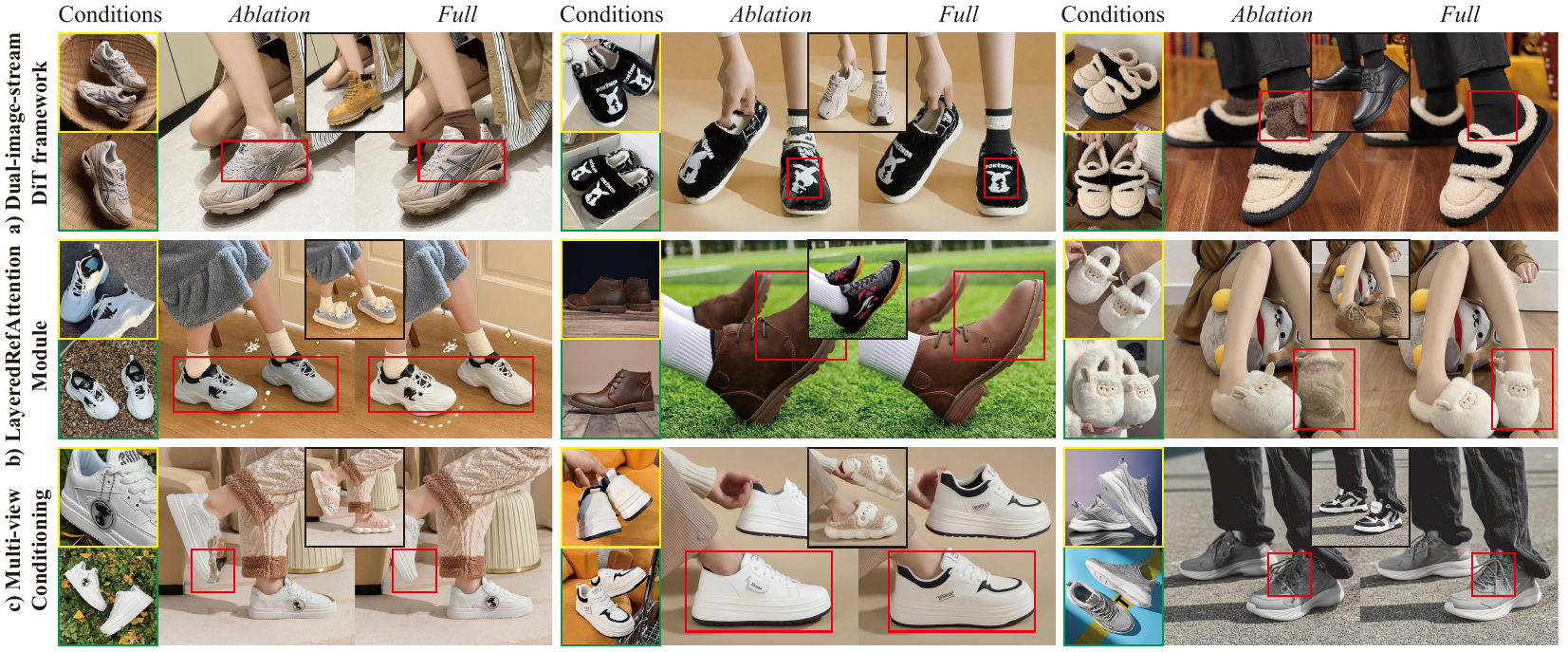}
%\vspace{-4mm}
  \caption{Ablation study. We highlight the improvements by all three components of method in red.} 
  %\vspace{-4mm}
  \label{fig: ablation}
\end{figure}

\subsection{Experimental Setup}

\noindent \textbf{Implementation Details.}
We employ the FLUX.1-Lite~\cite{flux} as the backbone and customize a dual-image-stream architecture as in Sec.~\ref{sec: dual-stream}. Our experiments are carried out on MVShoes at a resolution of $768 \times 768$, with 6305 pairs for training and 1000 pairs for testing. 
The model is trained for $7$ days on $8$ 80GB-A100 GPUs with DeepSpeed~\cite{deepspeed} ZeRO-2 to reduce memory usage, at a batch size of $4$. 
All parameters of DiT are trainable, using an AdamW optimizer~\cite{loshchilov2018adamw} with a constant learning rate of $3e-5$. 
At inference time, we run ShoeFit on a single A10 GPU for $25$ steps. The data augmentation follows the same protocol as in StableVTON~\cite{kim2023stableviton}. Please refer to Appendix for more details.

\noindent \textbf{Evaluation Protocols.}
Following evaluation in prior VTON works~\cite{morelli2023ladi}, we measure reconstruction accuracy by LPIPS~\cite{zhang2018perceptual}, SSIM~\cite{wang2004ssim} and DISTS~\cite{dists} in a paired setting given ground truth images, and authenticity of unpaired synthesized images by FID~\cite{parmar2021cleanfid} and KID~\cite{sutherland2018kid} without ground truth.

\noindent \textbf{Baselines.}
As a subtask of inpainting, we first select two widely used inpainting FLUX variants as baselines: FLUX.1-Fill~\cite{flux-fill} and FLUX.1-Fill-Redux~\cite{flux-redux}. The former involves concatenating product images with human images on the spatial dimension, similar to CatVTON~\cite{chong2024catvton}. The latter enhances the conditioning stream with SigLip~\cite{zhai2023sigmoidlosslanguageimage} to incorporate product shoe images. Additionally, we compare ShoeFit to two state-of-the-art methods in the VTON field: OOTDiffusion~\cite{xu2024ootdiffusion} based on the SDXL~\cite{podell2023sdxl} and CatVTON~\cite{chong2024catvton} built upon the FLUX.1-Lite. All models are implemented as originally deployed for single-view product shoe conditional generation. Considering CatVTON's capability to easily extend to multi-image injection, we further train a version of CatVTON to simultaneously inject two product views, the same as ShoeFit, denoted as CatVTON-FLUX-Lite-2V. Please refer to Appendix for more implementation details.

\subsection{Qualitative Results}
Fig.~\ref{fig: main} provides a qualitative comparison between ShoeFit and the baselines on MVShoes, addressing challenges such as complex structures, background interference, and difficult target poses. For a fair comparison, we report the results of the single-view conditioning version of all methods. Notably, we observe that OOTDiffusion~\cite{xu2024ootdiffusion} tends to exhibit extreme color variations, while FLUX.1-Fill-Redux~\cite{flux-redux} and FLUX.1-Fill~\cite{flux-fill}, possibly affected by the limitation of only LoRA~\cite{hu2022lora} being trainable (r=64), often produce visually plausible results that do not strictly adhere to the input conditions. In contrast, ShoeFit accurately generates structural details of the shoes (rows 1 and 2), decouples environmental lighting (row 3), avoids background confusion (row 4), and demonstrates robustness to uncommon poses (row 5). Please refer to the Appendix for more results.

\subsection{Quantitative Results}
For a fair comparison, we report the results of one/two product view conditioning generation separately in Tab.~\ref{tab: main}. FLUX.1-Fill-Redux~\cite{flux-redux} and FLUX.1-Fill~\cite{flux-fill} achieve the worst FID and DISTS scores, while OOTDiffusion~\cite{xu2024ootdiffusion} and CatVTON~\cite{chong2024catvton} show complementary performance in unpaired and paired evaluations. ShoesFit, benefiting from the additional specialized parameters provided by the Dual-image-stream DiT framework and background decoupling enabled by the LayeredRefAttention, significantly surpasses all baselines under both single-view and multi-view conditional generation.

\subsection{Ablation Study}
To validate our technical contributions, we define a Vanilla Model that uses FLUX.1-Lite as the backbone, deploying both a text-stream and an image-stream as in Fig.~\ref{fig: framework} (a). Multi-view Conditioning and LayeredRefAttention Module are removed, using only one footwear image as conditioning.

\noindent \textbf{Dual-image-stream DiT Framework.}
Since the text-stream only provides high-level semantics, its utility is limited in VFTON, which demands high image fidelity. Therefore, we first remove the text-stream and replicate a portion of the denoising stream’s architecture and weights to initialize the conditioning image stream, effectively decoupling the conditional feature extraction from the denoising process, while keeping the Vanilla Model’s other settings unchanged(denoted as ``+ D-i-s''). As shown in Tab.~\ref{tab: ablation} and Fig.~\ref{fig: ablation}, the vanilla model often generates artifacts and incorrect structures facing high-frequency textures due to the limitation in effective network capacity and the weakness in coupling of conditional images and noise. This demonstrates that the Dual-image-stream DiT framework plays a fundamental role in enhancing the model’s expressive capabilities.

\noindent \textbf{LayeredRefAttention Module.}
Observing the visual contamination from background and environmental lighting, we further integrate the LayeredRefAttention into the ``+ D-i-s'' version, denoted as ``+ D-i-s + LRA'', which effectively decouples the intrinsic appearance of the shoe from its surroundings. As illustrated in Tab.~\ref{tab: ablation} and Fig.~\ref{fig: ablation}, the model without LayeredRefAttention often fails to accurately reproduce the natural colors of the shoe where environmental lighting or shadows of product images are prominent, resulting in outputs that are mixed with environmental light or shadows. Additionally, mismatches with background patches in Attention can introduce colors confusion from the background content, such as slippers being contaminated by floor colors in Column 3 of Fig.~\ref{fig: ablation}. In contrast, the one with "LRA" effectively avoids such color contamination, delivering superior try-on results.

To investigate the efficacy of simple background masking, we conducted an ablation study utilizing standard attention while removing the background from the product shoe images. Visually, this naive approach exacerbates color artifacts, particularly for highly saturated hues. We hypothesize that this occurs because removing the background discards crucial illumination cues embedded in the product image, which are essential for the model to infer the shoe's intrinsic material properties and texture. Given that the training set contains inherent color variations between in-studio product shots and on-foot images due to disparate lighting, forcing the model to operate without background context introduces lighting ambiguity. Consequently, the model struggles to maintain color constancy, leading to degraded try-on performance and reduced visual fidelity.

\noindent \textbf{Multi-view Conditioning System.}
While a single source product view often fails to capture the complete texture of footwear, try-on results can benefit from multiple source views, thereby effectively reducing uncertainty in generation. Thus, we introduce the Multi-view Conditioning pipeline using Reference Concatenation and 3D RoPE in the ``+ D-i-s + LRA'' version, denoted as ``+ D-i-s +  LRA + MC''. As demonstrated in Tab.~\ref{tab: ablation} and Fig.~\ref{fig: ablation}, the model without ``MC'', which uses only one product shoe image as input (yellow box), often generates unrealistic speculations and artifacts in unseen areas due to insufficient conditional information. In contrast, ShoeFit benefits from the Multi-view Conditioning system, producing details that are more faithful to the input across different views. Furthermore, we also conduct an ablation study upon 3D RoPE's impact in Tab.~\ref{tab: ablation}, demonstrating the performance gain we achieve using the proposed RoPE without incurring any additional computational overhead. Please refer to the Appendix for more ablation results.

\begin{table}[t]
\caption{Quantitative ablation study of each component in Section of Method.} 
\centering
\begin{tabular}{lcccccc}
\toprule
\multicolumn{2}{l}{Method}          & SSIM $\uparrow$ & LPIPS $\downarrow$ & FID $\downarrow$ & KID $\downarrow$  & DISTS $\downarrow$ \\ 
\cmidrule{1-7}             
\multicolumn{2}{l}{Vanilla Model}    & 0.745 & 0.185  & 22.71  & 3.327 & 0.1179     \\
\multicolumn{2}{l}{+ D-i-s}         & 0.761 & 0.174  & 21.57  & 1.493 & 0.1038     \\
\multicolumn{2}{l}{+ D-i-s + Standard Attn (w/ mask)} & 0.756 & 0.181 & 21.70 & 1.698 & 0.1056 \\
\multicolumn{2}{l}{+ D-i-s + LRA (w/ mask)}     & 0.772 & 0.159  & 21.07  & 0.944 & 0.1059     \\ 
\midrule
\multicolumn{2}{l}{+ D-i-s + LRA + MC (w/o 3D RoPE)}     & 0.778 & 0.152  & 20.18  & 0.695 & 0.0971     \\ 
\multicolumn{2}{l}{+ D-i-s + LRA + MC \textbf{(Full)}}   & \textbf{0.780} & \textbf{0.149}  & \textbf{20.18}  & \textbf{0.485}  & \textbf{0.0941}    \\ 
\bottomrule
\end{tabular}

\label{tab: ablation}
\end{table}

\subsection{Inference Time and Memory Usage}
We evaluated the efficiency of our proposed enhancements by measuring inference time and memory footprint on an NVIDIA H20 GPU (with batch size of 1, 25 steps, and torch.bfloat16 precision). The results, presented in Table~\ref{tab: time cost}, reveal that the ShoeFit model with our Dual-image-stream (D-i-s) architecture substantially outperforms CAT-VTON in both speed and memory efficiency. This stems from removing the text stream and its computationally expensive T5 and CLIP encoders.

\begin{table}[!ht] 
    \centering % 使表格居中
    \caption{Comparison of inference time and GPU memory usage across different architectures.}
    \label{tab:inference_comparison} % 添加标签，方便在文本中交叉引用，例如 \ref{tab:inference_comparison}
    \begin{tabular}{l c c}
        \toprule
        \textbf{Method} & \textbf{Inference Time per Image} & \textbf{GPU Memory Usage} \\
        \midrule
        Cat-VTON (image + text streams) & 22.27 seconds & 35.5 GB \\
        ShoeFit (Dual-image-stream, D-i-s) & 17.44 seconds & 22.0 GB \\
        ShoeFit + D-i-s + LayeredRefAttention & 21.15 seconds & 28.5 GB \\
        \bottomrule
    \end{tabular}
\label{tab: time cost}
\end{table}

\section{Conclusion}
In conclusion, we formally define the VFTON problem and curate the first shoe try-on dataset, MVShoes, containing 7305 pairs of high-quality multi-view garment-model samples. Our proposed framework addresses the critical challenges of viewpoint misalignment and background-induced color distortion by leveraging a dual-stream architecture with Multi-View Conditioning and LayeredRefAttention modules. Through a comprehensive evaluation on the proposed MVShoes dataset, we demonstrate significant improvements in detail fidelity and texture consistency across diverse scenarios. This indicates that our method effectively mitigates background interference while preserving fine-grained shoe details across views, establishing a strong baseline for future VFTON research.

\noindent \textbf{Broader impacts.}
With the ability to synthesize images, arises the risk that ShoeFit might be used for inappropriate purposes such as producing media that breaches intellectual property rights or privacy norms. Because of these risks, we strongly advocate for the conscientious use of this technology.

\noindent\textbf{Limitation and future work.}
ShoeFit demonstrates substantial improvements in generating high-fidelity footwear try-on images across diverse scenes. However, certain limitations persist. Firstly, the model occasionally struggles with accurately rendering small logos and intricate text, particularly critical for e-commerce sellers. Secondly, we hope to introduce explicit 3D geometric and material priors in the future to achieve more robust multi-view representation and refined visual fidelity.

\clearpage

\section*{Acknowledgements} 
This work is supported by the Science and Technology Commission of Shanghai Municipality under research grant No. 25ZR1401187.

\bibliographystyle{splncs04}
\bibliography{main}

@String(ICLR  = {Int. Conf. Learn. Represent.})

@String(AAAI  = {AAAI})

@String(ICLR  = {ICLR})

@inproceedings{han2019clothflow,
  title={Clothflow: A flow-based model for clothed person generation},
  author={Han, Xintong and Hu, Xiaojun and Huang, Weilin and Scott, Matthew R},
  booktitle={Proceedings of the IEEE/CVF international conference on computer vision},
  year={2019}
}

@inproceedings{yang2020acgpn,
  title={Towards photo-realistic virtual try-on by adaptively generating-preserving image content},
  author={Yang, Han and Zhang, Ruimao and Guo, Xiaobao and Liu, Wei and Zuo, Wangmeng and Luo, Ping},
  booktitle={Proceedings of the IEEE/CVF conference on computer vision and pattern recognition},
  year={2020}
}

@inproceedings{Ge2021PFAFN,
  title={Parser-free virtual try-on via distilling appearance flows},
  author={Ge, Yuying and Song, Yibing and Zhang, Ruimao and Ge, Chongjian and Liu, Wei and Luo, Ping},
  booktitle={Proceedings of the IEEE/CVF conference on computer vision and pattern recognition},
  year={2021}
}

@inproceedings{lee2022hrviton,
  title={High-resolution virtual try-on with misalignment and occlusion-handled conditions},
  author={Lee, Sangyun and Gu, Gyojung and Park, Sunghyun and Choi, Seunghwan and Choo, Jaegul},
  booktitle={European Conference on Computer Vision},
  year={2022},
}

@inproceedings{morelli2022dresscode,
  title={Dress code: High-resolution multi-category virtual try-on},
  author={Davide, Morelli and Matteo, Fincato and Marcella, Cornia and Federico, Landi and Fabio, Cesari and Rita, Cucchiara},
  booktitle={Proceedings of the IEEE/CVF Conference on Computer Vision and Pattern Recognition},
  year={2022}
}

@inproceedings{bai2022sdafn,
  title={Single stage virtual try-on via deformable attention flows},
  author={Bai, Shuai and Zhou, Huiling and Li, Zhikang and Zhou, Chang and Yang, Hongxia},
  booktitle={European Conference on Computer Vision},
  year={2022},
}

@inproceedings{han2018viton,
  title={Viton: An image-based virtual try-on network},
  author={Han, Xintong and Wu, Zuxuan and Wu, Zhe and Yu, Ruichi and Davis, Larry S},
  booktitle={Proceedings of the IEEE conference on computer vision and pattern recognition},
  year={2018}
}

@article{goodfellow2014GAN,
  title={Generative adversarial nets},
  author={Goodfellow, Ian and Pouget-Abadie, Jean and Mirza, Mehdi and Xu, Bing and Warde-Farley, David and Ozair, Sherjil and Courville, Aaron and Bengio, Yoshua},
  journal={Advances in neural information processing systems},
  year={2014}
}

@article{bookstein1989tps,
  title={Principal warps: Thin-plate splines and the decomposition of deformations},
  author={Bookstein, Fred L.},
  journal={IEEE Transactions on pattern analysis and machine intelligence},
  year={1989},
}

@article{huang2023composer,
  title={Composer: Creative and controllable image synthesis with composable conditions},
  author={Huang, Lianghua and Chen, Di and Liu, Yu and Shen, Yujun and Zhao, Deli and Zhou, Jingren},
  journal={arXiv preprint arXiv:2302.09778},
  year={2023}
}

@article{chen2023anydoor,
  title={Anydoor: Zero-shot object-level image customization},
  author={Chen, Xi and Huang, Lianghua and Liu, Yu and Shen, Yujun and Zhao, Deli and Zhao, Hengshuang},
  journal={arXiv preprint arXiv:2307.09481},
  year={2023}
}

@misc{deepspeed,
    title = {DeepSpeed},
key = {40},
    howpublished = {\url{https://github.com/microsoft/DeepSpeed}}
}

@article{nichol2021glide,
  title={Glide: Towards photorealistic image generation and editing with text-guided diffusion models},
  author={Nichol, Alex and Dhariwal, Prafulla and Ramesh, Aditya and Shyam, Pranav and Mishkin, Pamela and McGrew, Bob and Sutskever, Ilya and Chen, Mark},
  journal={arXiv preprint arXiv:2112.10741},
  year={2021}
}

@inproceedings{rombach2022ldm,
  title={High-resolution image synthesis with latent diffusion models},
  author={Rombach, Robin and Blattmann, Andreas and Lorenz, Dominik and Esser, Patrick and Ommer, Bj{\"o}rn},
  booktitle={Proceedings of the IEEE/CVF conference on computer vision and pattern recognition},
  year={2022}
}

@inproceedings{lin2017fpn,
  title={Feature pyramid networks for object detection},
  author={Lin, Tsung-Yi and Doll{\'a}r, Piotr and Girshick, Ross and He, Kaiming and Hariharan, Bharath and Belongie, Serge},
  booktitle={Proceedings of the IEEE conference on computer vision and pattern recognition},
  year={2017}
}

@inproceedings{radford2021clip,
  title={Learning transferable visual models from natural language supervision},
  author={Radford, Alec and Kim, Jong Wook and Hallacy, Chris and Ramesh, Aditya and Goh, Gabriel and Agarwal, Sandhini and Sastry, Girish and Askell, Amanda and Mishkin, Pamela and Clark, Jack and others},
  booktitle={International conference on machine learning},
  year={2021},
}

@inproceedings{gou2023dci,
  title={Taming the Power of Diffusion Models for High-Quality Virtual Try-On with Appearance Flow},
  author={Gou, Junhong and Sun, Siyu and Zhang, Jianfu and Si, Jianlou and Qian, Chen and Zhang, Liqing},
  booktitle={Proceedings of the 31st ACM International Conference on Multimedia},
  year={2023}
}

@article{wang2004ssim,
  title={Image quality assessment: from error visibility to structural similarity},
  author={Wang, Zhou and Bovik, Alan C and Sheikh, Hamid R and Simoncelli, Eero P},
  journal={IEEE transactions on image processing},
  year={2004},
}

@inproceedings{parmar2021cleanfid,
  title={On aliased resizing and surprising subtleties in gan evaluation},
  author={Parmar, Gaurav and Zhang, Richard and Zhu, Jun-Yan},
  booktitle={Proceedings of the IEEE/CVF Conference on Computer Vision and Pattern Recognition},
  year={2022}
}

@inproceedings{zhang2018perceptual,
  title={The unreasonable effectiveness of deep features as a perceptual metric},
  author={Zhang, Richard and Isola, Phillip and Efros, Alexei A and Shechtman, Eli and Wang, Oliver},
  booktitle={Proceedings of the IEEE conference on computer vision and pattern recognition},
  year={2018}
}

@inproceedings{chopra2021zflow,
  title={Zflow: Gated appearance flow-based virtual try-on with 3d priors},
  author={Chopra, Ayush and Jain, Rishabh and Hemani, Mayur and Krishnamurthy, Balaji},
  booktitle={Proceedings of the IEEE/CVF International Conference on Computer Vision},
  year={2021}
}

@inproceedings{morelli2023ladi,
  title={{LaDI-VTON: Latent Diffusion Textual-Inversion Enhanced Virtual Try-On}},
  author={Morelli, Davide and Baldrati, Alberto and Cartella, Giuseppe and Cornia, Marcella and Bertini, Marco and Cucchiara, Rita},
  booktitle={Proceedings of the ACM International Conference on Multimedia},
  year={2023}
}

@inproceedings{kim2023stableviton,
    author = {Kim, Jeongho and Gu, Gyojung and Park, Minho and Park, Sunghyun and Choo, Jaegul},
    title = {StableVITON: Learning Semantic Correspondence with Latent Diffusion Model for Virtual Try-On},
    booktitle = {Proceedings of the IEEE/CVF Conference on Computer Vision and Pattern Recognition},
    year = {2024} 
}

@article{kirillov2023sam,
  title={Segment anything},
  author={Kirillov, Alexander and Mintun, Eric and Ravi, Nikhila and Mao, Hanzi and Rolland, Chloe and Gustafson, Laura and Xiao, Tete and Whitehead, Spencer and Berg, Alexander C and Lo, Wan-Yen and others},
  journal={arXiv preprint arXiv:2304.02643},
  year={2023}
}

@misc{oquab2023dinov2,
  title={DINOv2: Learning Robust Visual Features without Supervision},
  author={Oquab, Maxime and Darcet, Timothée and Moutakanni, Theo and Vo, Huy V. and Szafraniec, Marc and Khalidov, Vasil and Fernandez, Pierre and Haziza, Daniel and Massa, Francisco and El-Nouby, Alaaeldin and Howes, Russell and Huang, Po-Yao and Xu, Hu and Sharma, Vasu and Li, Shang-Wen and Galuba, Wojciech and Rabbat, Mike and Assran, Mido and Ballas, Nicolas and Synnaeve, Gabriel and Misra, Ishan and Jegou, Herve and Mairal, Julien and Labatut, Patrick and Joulin, Armand and Bojanowski, Piotr},
  journal={arXiv:2304.07193},
  year={2023}
}

@article{kingma2013vae,
  title={Auto-encoding variational bayes},
  author={Kingma, Diederik P and Welling, Max},
  journal={arXiv preprint arXiv:1312.6114},
  year={2013}
}

@article{ho2020ddpm,
  title={Denoising diffusion probabilistic models},
  author={Ho, Jonathan and Jain, Ajay and Abbeel, Pieter},
  journal={Advances in neural information processing systems},
  year={2020}
}

@article{achiam2023gpt,
  title={Gpt-4 technical report},
  author={Achiam, Josh and Adler, Steven and Agarwal, Sandhini and Ahmad, Lama and Akkaya, Ilge and Aleman, Florencia Leoni and Almeida, Diogo and Altenschmidt, Janko and Altman, Sam and Anadkat, Shyamal and others},
  journal={arXiv preprint arXiv:2303.08774},
  year={2023}
}

@article{loshchilov2018adamw,
  title={Fixing weight decay regularization in adam},
  author={Loshchilov, Ilya and Hutter, Frank},
  year={2018}
}

@inproceedings{sutherland2018kid,
  title={Demystifying MMD GANs},
  author={Sutherland, JD and Arbel, Michael and Gretton, Arthur},
  booktitle={International Conference for Learning Representations},
  year={2018}
}

@article{podell2023sdxl,
  title={Sdxl: Improving latent diffusion models for high-resolution image synthesis},
  author={Podell, Dustin and English, Zion and Lacey, Kyle and Blattmann, Andreas and Dockhorn, Tim and M{\"u}ller, Jonas and Penna, Joe and Rombach, Robin},
  journal={arXiv preprint arXiv:2307.01952},
  year={2023}
}

@article{choi2024idm,
  title={Improving Diffusion Models for Virtual Try-on},
  author={Choi, Yisol and Kwak, Sangkyung and Lee, Kyungmin and Choi, Hyungwon and Shin, Jinwoo},
  journal={arXiv preprint arXiv:2403.05139},
  year={2024}
}

@article{chen2024wearanyway,
      title={Wear-Any-Way: Manipulable Virtual Try-on via Sparse Correspondence Alignment},
      author={Chen, Mengting and Chen, Xi and Zhai, Zhonghua and Ju, Chen and Hong, Xuewen and Lan, Jinsong and Xiao, Shuai},
      journal={arXiv preprint},
      year={2024}
    }

@article{xu2024ootdiffusion,
  title={OOTDiffusion: Outfitting Fusion based Latent Diffusion for Controllable Virtual Try-on},
  author={Xu, Yuhao and Gu, Tao and Chen, Weifeng and Chen, Chengcai},
  journal={arXiv preprint arXiv:2403.01779},
  year={2024}
}

@article{zhang2024mmtryon,
  title={MMTryon: Multi-Modal Multi-Reference Control for High-Quality Fashion Generation},
  author={Zhang, Xujie and Lin, Ente and Li, Xiu and Luo, Yuxuan and Kampffmeyer, Michael and Dong, Xin and Liang, Xiaodan},
  journal={arXiv preprint arXiv:2405.00448},
  year={2024}
}

@inproceedings{dong2020fashion,
  title={Fashion editing with adversarial parsing learning},
  author={Dong, Haoye and Liang, Xiaodan and Zhang, Yixuan and Zhang, Xujie and Shen, Xiaohui and Xie, Zhenyu and Wu, Bowen and Yin, Jian},
  booktitle={Proceedings of the IEEE/CVF conference on computer vision and pattern recognition},
  pages={8120--8128},
  year={2020}
}

@inproceedings{jo2019scfegan,
  title={Sc-fegan: Face editing generative adversarial network with user's sketch and color},
  author={Jo, Youngjoo and Park, Jongyoul},
  booktitle={Proceedings of the IEEE/CVF international conference on computer vision},
  pages={1745--1753},
  year={2019}
}

@inproceedings{lee2020maskgan,
  title={Maskgan: Towards diverse and interactive facial image manipulation},
  author={Lee, Cheng-Han and Liu, Ziwei and Wu, Lingyun and Luo, Ping},
  booktitle={Proceedings of the IEEE/CVF conference on computer vision and pattern recognition},
  pages={5549--5558},
  year={2020}
}

@article{ye2023ipadapter,
  title={Ip-adapter: Text compatible image prompt adapter for text-to-image diffusion models},
  author={Ye, Hu and Zhang, Jun and Liu, Sibo and Han, Xiao and Yang, Wei},
  journal={arXiv preprint arXiv:2308.06721},
  year={2023}
}

@inproceedings{zhang2023controlnet,
  title={Adding conditional control to text-to-image diffusion models},
  author={Zhang, Lvmin and Rao, Anyi and Agrawala, Maneesh},
  booktitle={Proceedings of the IEEE/CVF International Conference on Computer Vision},
  pages={3836--3847},
  year={2023}
}

@article{chong2024catvton,
  title={Catvton: Concatenation is all you need for virtual try-on with diffusion models},
  author={Chong, Zheng and Dong, Xiao and Li, Haoxiang and Zhang, Shiyue and Zhang, Wenqing and Zhang, Xujie and Zhao, Hanqing and Jiang, Dongmei and Liang, Xiaodan},
  journal={arXiv preprint arXiv:2407.15886},
  year={2024}
}

@inproceedings{li2023gligen,
  title={Gligen: Open-set grounded text-to-image generation},
  author={Li, Yuheng and Liu, Haotian and Wu, Qingyang and Mu, Fangzhou and Yang, Jianwei and Gao, Jianfeng and Li, Chunyuan and Lee, Yong Jae},
  booktitle={Proceedings of the IEEE/CVF Conference on Computer Vision and Pattern Recognition},
  pages={22511--22521},
  year={2023}
}

@inproceedings{yang2023reco,
  title={Reco: Region-controlled text-to-image generation},
  author={Yang, Zhengyuan and Wang, Jianfeng and Gan, Zhe and Li, Linjie and Lin, Kevin and Wu, Chenfei and Duan, Nan and Liu, Zicheng and Liu, Ce and Zeng, Michael and others},
  booktitle={Proceedings of the IEEE/CVF Conference on Computer Vision and Pattern Recognition},
  pages={14246--14255},
  year={2023}
}

@article{zhang2023controllablegpt4,
  title={Controllable text-to-image generation with gpt-4},
  author={Zhang, Tianjun and Zhang, Yi and Vineet, Vibhav and Joshi, Neel and Wang, Xin},
  journal={arXiv preprint arXiv:2305.18583},
  year={2023}
}

@inproceedings{chen2024layoutcontrol,
  title={Training-free layout control with cross-attention guidance},
  author={Chen, Minghao and Laina, Iro and Vedaldi, Andrea},
  booktitle={Proceedings of the IEEE/CVF Winter Conference on Applications of Computer Vision},
  pages={5343--5353},
  year={2024}
}

@inproceedings{ju2023humansd,
  title={Humansd: A native skeleton-guided diffusion model for human image generation},
  author={Ju, Xuan and Zeng, Ailing and Zhao, Chenchen and Wang, Jianan and Zhang, Lei and Xu, Qiang},
  booktitle={Proceedings of the IEEE/CVF International Conference on Computer Vision},
  pages={15988--15998},
  year={2023}
}

@inproceedings{xie2023boxdiff,
  title={Boxdiff: Text-to-image synthesis with training-free box-constrained diffusion},
  author={Xie, Jinheng and Li, Yuexiang and Huang, Yawen and Liu, Haozhe and Zhang, Wentian and Zheng, Yefeng and Shou, Mike Zheng},
  booktitle={Proceedings of the IEEE/CVF International Conference on Computer Vision},
  pages={7452--7461},
  year={2023}
}

@inproceedings{mou2024t2iadapter,
  title={T2i-adapter: Learning adapters to dig out more controllable ability for text-to-image diffusion models},
  author={Mou, Chong and Wang, Xintao and Xie, Liangbin and Wu, Yanze and Zhang, Jian and Qi, Zhongang and Shan, Ying},
  booktitle={Proceedings of the AAAI Conference on Artificial Intelligence},
  volume={38},
  number={5},
  pages={4296--4304},
  year={2024}
}

@misc{flux,
    title = {FLUX.1-dev},
    howpublished = {\url{https://github.com/black-forest-labs/flux}},
    author = {Black-forest-labs},
    year = {2024},
    url = {https://github.com/black-forest-labs/flux},
}

@misc{nike,
    title = {NIKE, INC. ANNUAL REPORT},
    howpublished = {\url{https://www.sec.gov/Archives/edgar/data/320187/000032018724000044/nke-20240531.htm}},
    author = {NIKE, INC.},
    year = {2024},
    url = {https://www.sec.gov/Archives/edgar/data/320187/000032018724000044/nke-20240531.htm},
}

@inproceedings{sd3,
  title={Scaling rectified flow transformers for high-resolution image synthesis},
  author={Esser, Patrick and Kulal, Sumith and Blattmann, Andreas and Entezari, Rahim and M{\"u}ller, Jonas and Saini, Harry and Levi, Yam and Lorenz, Dominik and Sauer, Axel and Boesel, Frederic and others},
  booktitle={Forty-first international conference on machine learning},
  year={2024}
}

@misc{flux-fill,
    title = {FLUX.1-Fill-dev},
    howpublished = {\url{https://huggingface.co/black-forest-labs/FLUX.1-Fill-dev}},
    author = {Black-forest-labs},
    year = {2024},
    url = {https://huggingface.co/black-forest-labs/FLUX.1-Fill-dev},
}

@misc{flux-redux,
    title = {FLUX.1-Redux-dev},
    howpublished = {\url{https://huggingface.co/black-forest-labs/FLUX.1-Redux-dev}},
    author = {Black-forest-labs},
    year = {2024},
    url = {https://huggingface.co/black-forest-labs/FLUX.1-Redux-dev},
}

@misc{3D_projection,
    title = {3D Projection},
    howpublished = {\url{https://en.wikipedia.org/wiki/3D_projection}},
    author = {Wikipedia},
    year = {2025},
    url = {https://en.wikipedia.org/wiki/3D_projection},
}

@inproceedings{dit,
  title={Scalable diffusion models with transformers},
  author={Peebles, William and Xie, Saining},
  booktitle={Proceedings of the IEEE/CVF international conference on computer vision},
  pages={4195--4205},
  year={2023}
}

@article{li2024anyfit,
  title={AnyFit: Controllable Virtual Try-on for Any Combination of Attire Across Any Scenario},
  author={Li, Yuhan and Zhou, Hao and Shang, Wenxiang and Lin, Ran and Chen, Xuanhong and Ni, Bingbing},
  journal={arXiv preprint arXiv:2405.18172},
  year={2024}
}

@article{zhang2024boow,
  title={BooW-VTON: Boosting In-the-Wild Virtual Try-On via Mask-Free Pseudo Data Training},
  author={Zhang, Xuanpu and Song, Dan and Zhan, Pengxin and Chen, Qingguo and Xu, Zhao and Luo, Weihua and Zhang, Kaifu and Liu, Anan},
  journal={arXiv preprint arXiv:2408.06047},
  year={2024}
}

@article{tan2024ominicontrol,
  title={Ominicontrol: Minimal and universal control for diffusion transformer},
  author={Tan, Zhenxiong and Liu, Songhua and Yang, Xingyi and Xue, Qiaochu and Wang, Xinchao},
  journal={arXiv preprint arXiv:2411.15098},
  year={2024}
}

@article{wang2025unicombine,
  title={UniCombine: Unified Multi-Conditional Combination with Diffusion Transformer},
  author={Wang, Haoxuan and Peng, Jinlong and He, Qingdong and Yang, Hao and Jin, Ying and Wu, Jiafu and Hu, Xiaobin and Pan, Yanjie and Gan, Zhenye and Chi, Mingmin and others},
  journal={arXiv preprint arXiv:2503.09277},
  year={2025}
}

@inproceedings{hu2018squeeze,
  title={Squeeze-and-excitation networks},
  author={Hu, Jie and Shen, Li and Sun, Gang},
  booktitle={Proceedings of the IEEE conference on computer vision and pattern recognition},
  pages={7132--7141},
  year={2018}
}

@article{dists,
  title={Image quality assessment: Unifying structure and texture similarity},
  author={Ding, Keyan and Ma, Kede and Wang, Shiqi and Simoncelli, Eero P},
  journal={IEEE transactions on pattern analysis and machine intelligence},
  volume={44},
  number={5},
  pages={2567--2581},
  year={2020},
  publisher={IEEE}
}

@article{shi2023mvdream,
  title={Mvdream: Multi-view diffusion for 3d generation},
  author={Shi, Yichun and Wang, Peng and Ye, Jianglong and Long, Mai and Li, Kejie and Yang, Xiao},
  journal={arXiv preprint arXiv:2308.16512},
  year={2023}
}

@article{liu2023one,
  title={One-2-3-45: Any single image to 3d mesh in 45 seconds without per-shape optimization},
  author={Liu, Minghua and Xu, Chao and Jin, Haian and Chen, Linghao and Varma T, Mukund and Xu, Zexiang and Su, Hao},
  journal={Advances in Neural Information Processing Systems},
  volume={36},
  pages={22226--22246},
  year={2023}
}

@article{shi2023zero123++,
  title={Zero123++: a single image to consistent multi-view diffusion base model},
  author={Shi, Ruoxi and Chen, Hansheng and Zhang, Zhuoyang and Liu, Minghua and Xu, Chao and Wei, Xinyue and Chen, Linghao and Zeng, Chong and Su, Hao},
  journal={arXiv preprint arXiv:2310.15110},
  year={2023}
}

@inproceedings{kocsis2024lightit,
  title={Lightit: Illumination modeling and control for diffusion models},
  author={Kocsis, Peter and Philip, Julien and Sunkavalli, Kalyan and Nie{\ss}ner, Matthias and Hold-Geoffroy, Yannick},
  booktitle={Proceedings of the IEEE/CVF Conference on Computer Vision and Pattern Recognition},
  pages={9359--9369},
  year={2024}
}

@inproceedings{karras2019style,
  title={A style-based generator architecture for generative adversarial networks},
  author={Karras, Tero and Laine, Samuli and Aila, Timo},
  booktitle={Proceedings of the IEEE/CVF conference on computer vision and pattern recognition},
  pages={4401--4410},
  year={2019}
}

@article{su2024roformer,
  title={Roformer: Enhanced transformer with rotary position embedding},
  author={Su, Jianlin and Ahmed, Murtadha and Lu, Yu and Pan, Shengfeng and Bo, Wen and Liu, Yunfeng},
  journal={Neurocomputing},
  volume={568},
  pages={127063},
  year={2024},
  publisher={Elsevier}
}

@article{hu2022lora,
  title={Lora: Low-rank adaptation of large language models.},
  author={Hu, Edward J and Shen, Yelong and Wallis, Phillip and Allen-Zhu, Zeyuan and Li, Yuanzhi and Wang, Shean and Wang, Lu and Chen, Weizhu and others},
  journal={ICLR},
  volume={1},
  number={2},
  pages={3},
  year={2022}
}

@misc{yang2023effectivewholebodyposeestimation,
      title={Effective Whole-body Pose Estimation with Two-stages Distillation}, 
      author={Zhendong Yang and Ailing Zeng and Chun Yuan and Yu Li},
      year={2023},
      eprint={2307.15880},
      archivePrefix={arXiv},
      primaryClass={cs.CV},
      url={https://arxiv.org/abs/2307.15880}, 
}

@misc{liu2024groundingdinomarryingdino,
      title={Grounding DINO: Marrying DINO with Grounded Pre-Training for Open-Set Object Detection}, 
      author={Shilong Liu and Zhaoyang Zeng and Tianhe Ren and Feng Li and Hao Zhang and Jie Yang and Qing Jiang and Chunyuan Li and Jianwei Yang and Hang Su and Jun Zhu and Lei Zhang},
      year={2024},
      eprint={2303.05499},
      archivePrefix={arXiv},
      primaryClass={cs.CV},
      url={https://arxiv.org/abs/2303.05499}, 
}

@misc{bai2023qwenvlversatilevisionlanguagemodel,
      title={Qwen-VL: A Versatile Vision-Language Model for Understanding, Localization, Text Reading, and Beyond}, 
      author={Jinze Bai and Shuai Bai and Shusheng Yang and Shijie Wang and Sinan Tan and Peng Wang and Junyang Lin and Chang Zhou and Jingren Zhou},
      year={2023},
      eprint={2308.12966},
      archivePrefix={arXiv},
      primaryClass={cs.CV},
      url={https://arxiv.org/abs/2308.12966}, 
}

@article{guo2025any2anytryon,
  title={Any2AnyTryon: Leveraging Adaptive Position Embeddings for Versatile Virtual Clothing Tasks},
  author={Guo, Hailong and Zeng, Bohan and Song, Yiren and Zhang, Wentao and Zhang, Chuang and Liu, Jiaming},
  journal={arXiv preprint arXiv:2501.15891},
  year={2025}
}

@misc{zhai2023sigmoidlosslanguageimage,
      title={Sigmoid Loss for Language Image Pre-Training}, 
      author={Xiaohua Zhai and Basil Mustafa and Alexander Kolesnikov and Lucas Beyer},
      year={2023},
      eprint={2303.15343},
      archivePrefix={arXiv},
      primaryClass={cs.CV},
      url={https://arxiv.org/abs/2303.15343}, 
}

@misc{raffel2023exploringlimitstransferlearning,
      title={Exploring the Limits of Transfer Learning with a Unified Text-to-Text Transformer}, 
      author={Colin Raffel and Noam Shazeer and Adam Roberts and Katherine Lee and Sharan Narang and Michael Matena and Yanqi Zhou and Wei Li and Peter J. Liu},
      year={2023},
      eprint={1910.10683},
      archivePrefix={arXiv},
      primaryClass={cs.LG},
      url={https://arxiv.org/abs/1910.10683}, 
}

@article{lipman2022flow,
  title={Flow matching for generative modeling},
  author={Lipman, Yaron and Chen, Ricky TQ and Ben-Hamu, Heli and Nickel, Maximilian and Le, Matt},
  journal={arXiv preprint arXiv:2210.02747},
  year={2022}
}

\appendix
\clearpage
\setcounter{page}{1}

\begin{itemize}

\item In Section~\ref{supp: sec: limit}, we provide a more detailed discussion about the limitations of ShoeFit. 
\item In Section~\ref{supp: sec: dataset}, we provide additional details about the MVShoes datasets, including data processing and data statistics.
\item In Section~\ref{supp: sec: supplement_method}, we provide preliminary knowledge about the FLUX~\cite{flux} model, as well as how LayeredRefAttention modules are applied in the Single Stream DiT Block.
\item In Section~\ref{supp: sec: imple_details}, we provide additional details about the experimental setup, including baseline training, data augmentation, and hyperparameters.
\item In Section~\ref{supp: sec: more_results}, we present a multitude of ShoeFit generated images, including more ablation results, more comparisons with the baselines, with additional results displayed in challenging scenarios.

\end{itemize}
%--------------------------------------------------------------------------

\begin{wrapfigure}{r}{0.5\textwidth}

  \centering
  \includegraphics[width=0.48\textwidth]{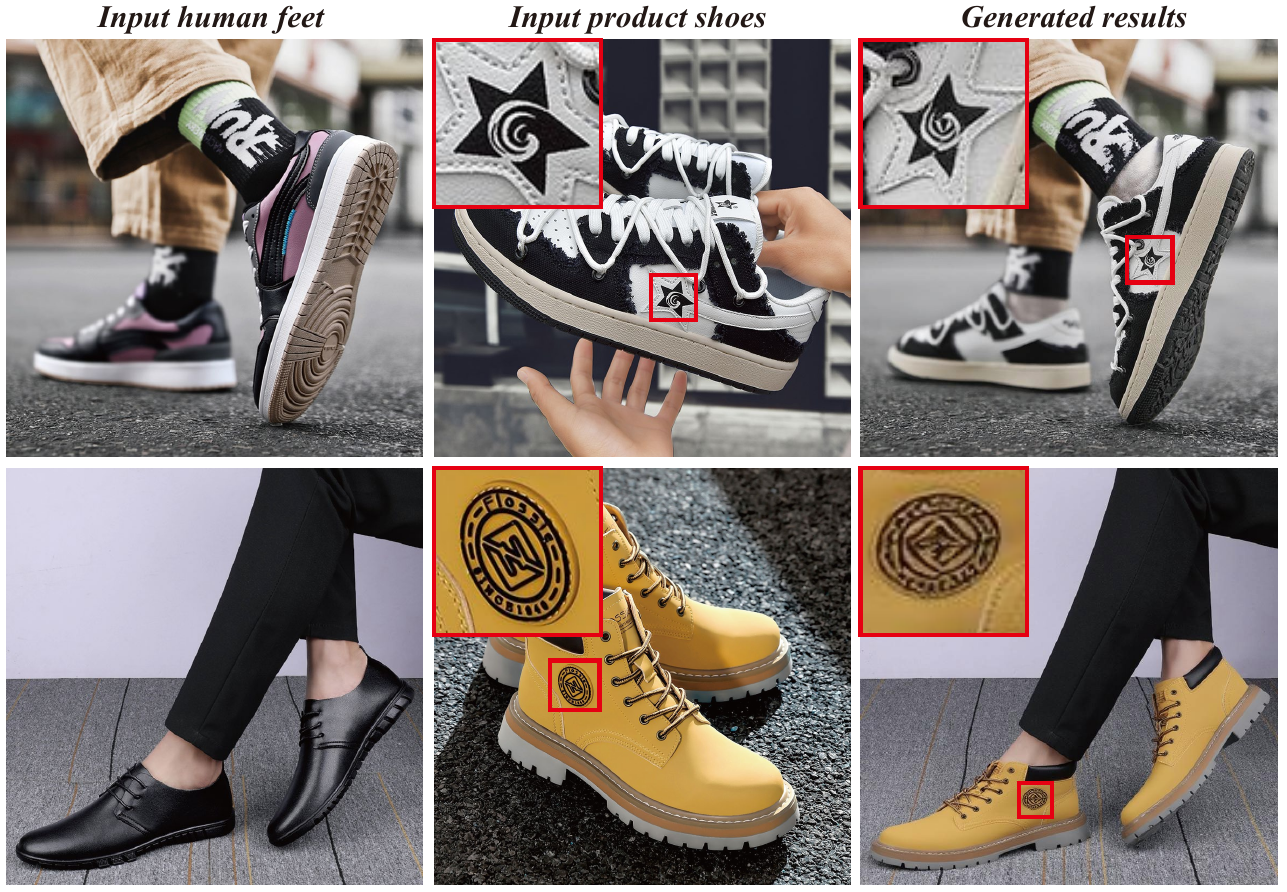}

  \caption{Similar to the common constraints faced by existing generative models, ShoeFit occasionally struggles to accurately render small logos and intricate text due to their small portion in images and high-frequency variations.} 
  \label{supp: fig: limit}

\end{wrapfigure}

\section{Limitations and Future Work}
\label{supp: sec: limit}
% LIMIATATION的展示
We propose ShoeFit, a dual-stream DiT framework that addresses the critical challenges of viewpoint misalignment and background-induced color distortion in VFTON by Multi-View Conditioning and LayeredRefAttention modules. 
However, certain limitations persist. Firstly, similar to the common constraints faced by existing generative models, the model occasionally struggles to accurately render small logos and intricate text due to their small portion in images and high-frequency variations. We illustrate this limitation in Fig.~\ref{supp: fig: limit}. These small logos are particularly critical for e-commerce sellers as they often convey brand information that consumers care about. Therefore, we plan to develop a detailed supplementary conditioning method in future research to pre-detect and enhance the injection of such patterns, fundamentally addressing this issue. Secondly, we aim to introduce explicit 3D geometric and material priors in the future to achieve more robust multi-view representation and refined visual fidelity.

%--------------------------------------------------------------------------

\section{MVShoes Dataset Supplement}
\label{supp: sec: dataset}
In the main body of the paper, we provide a brief description of the data processing pipeline and a rough visual presentation of the dataset statistics in terms of Rich Scenes and Diverse Footwear Categories due to space constraints. In this section, we offer more detailed information on the data processing pipeline to ensure reader comprehension. Additionally, we present the specific distribution statistics of MVShoes in a quantitative table format.

\paragraph{Data Processing Details.}  
\label{supp: sec: process}
Given a dataset comprising raw images of various shoes and human models, we employ Qwen2.5-VL~\cite{bai2023qwenvlversatilevisionlanguagemodel}, which operates with 7 billion parameters, to distinguish between footwear images and human model images. The prompting framework utilized is as follows: ``Analyze the provided image and determine whether it depicts a model image or a shoe image. A model image is defined as one that portrays a human model’s lower body or legs adorned with shoes, while a shoe image solely comprises images of shoes without any human presence. Assign a value of `1' if it qualifies as a model image and `0' if it does not. ''

Following this classification, we extract image features from the shoe images using the CLIP~\cite{radford2021clip} model, implementing a similarity threshold of 0.9 to effectively eliminate duplicate images within the dataset. Subsequently, we perform segmentation by SAM~\cite{kirillov2023sam} to isolate the shoe regions in each image, aligning the DINO~\cite{oquab2023dinov2} features to filter out shoes-model data pairs exhibiting inner similarities greater than 0.8.

This procedure facilitates the initial construction of human-footwear try-on triplets, which are then subjected to manual filtering to address any potential errors or oversights. Additionally, we exclude visually blurred images and compile statistical analyses pertaining to shoe categories and try-on scenarios. In accordance with methodologies established in DWPose~\cite{yang2023effectivewholebodyposeestimation} and GroundingDINO~\cite{liu2024groundingdinomarryingdino}, we extract foot poses and shoe masks, ultimately resulting in the generation of high-resolution, category-comprehensive try-on triplets. All models referenced above are open-source, and their respective URLs are provided as follows:

\begin{itemize}
    \item Qwen2.5-VL-7B~\cite{bai2023qwenvlversatilevisionlanguagemodel}: https://huggingface.co/Qwen/Qwen2.5-VL-7B-Instruct
    \item SAM~\cite{kirillov2023sam}: https://github.com/facebookresearch/segment-anything
    \item CLIP~\cite{radford2021clip}: https://github.com/openai/CLIP
    \item DINO~\cite{oquab2023dinov2}: https://github.com/facebookresearch/dinov2
    \item DWPose~\cite{yang2023effectivewholebodyposeestimation}: https://github.com/IDEA-Research/DWPose
    \item GroundingDINO~\cite{liu2024groundingdinomarryingdino}: https://github.com/IDEA-Research/GroundingDINO
\end{itemize}

\begin{figure}[t]
  \centering
\includegraphics[width=1.0\textwidth]{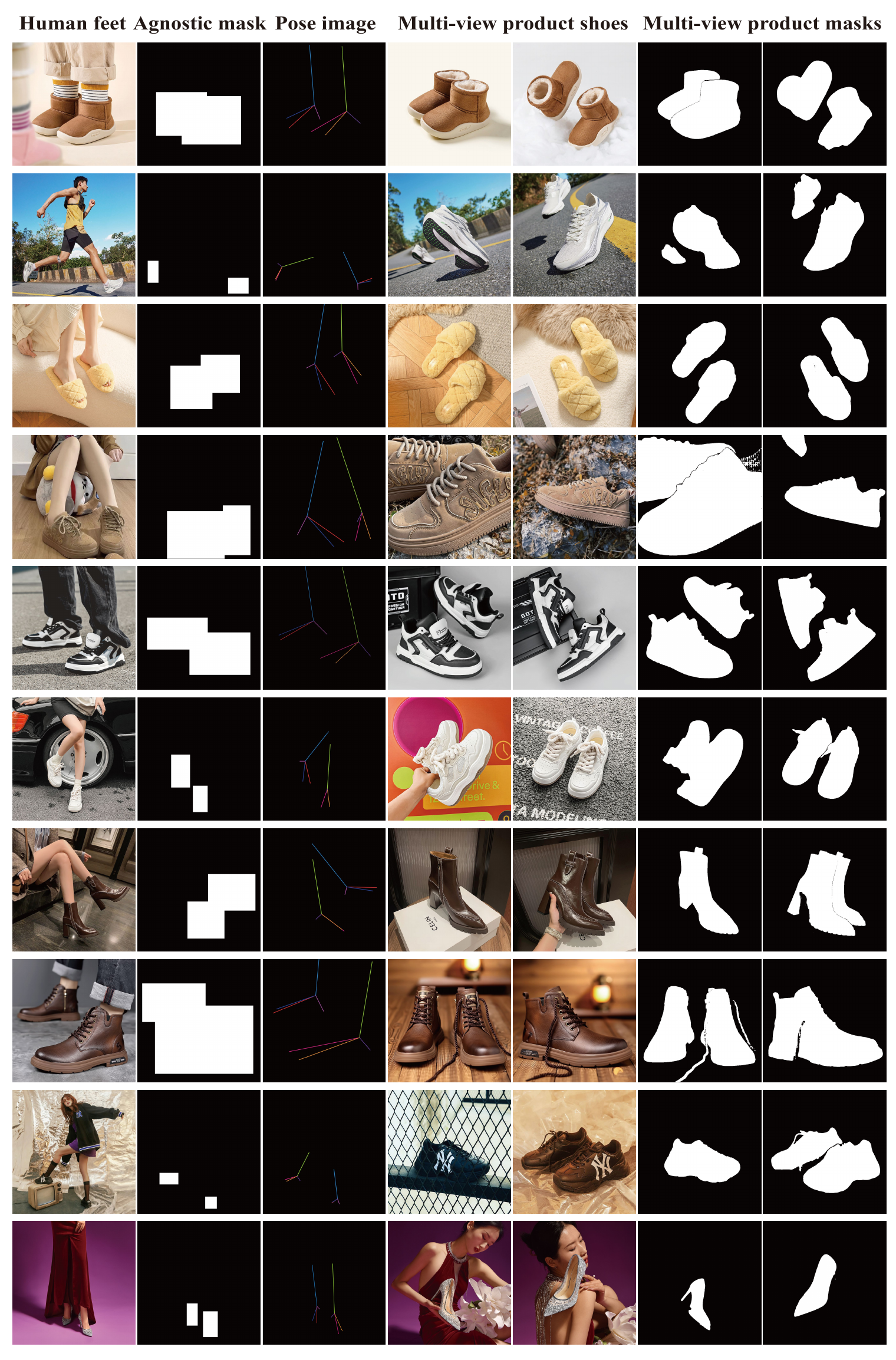}
  \caption{We provide more visual samples from MVShoes datasets.} 
  \label{supp: tab: datashow}
\end{figure}

\paragraph{Dataset Statistics.}  
The dataset features diverse shoe categories and rich human scenes. We present the quantitative distribution statistics of MVShoes in Tab.~\ref{supp: tab: cate} and Tab.~\ref{supp: tab: scene}. We also provide more visual samples from MVShoes datasets in Fig.~\ref{supp: tab: datashow}.

\paragraph{Pose Estimation for Shoe Try-On.}
Traditional open-source pose estimation methods (e.g., OpenPose, DWPose) in VITON pipeline, struggle to capture lower-body/foot poses in VFTON, which often only include partial leg shots or full lower-body views. To address this, we fine-tune the dwpose model on specially collected datasets containing diverse lower-body and close-up leg poses. This enables our model to generate accurate pose descriptions for various leg postures, significantly enhancing its generalization capability for shoe-related applications.

\paragraph{Generalizable Foot Mask Design.}
Given the diversity of shoe types (e.g., sneakers, boots, sandals) and their varying leg coverage, we propose a leg-pose-aware mask generation strategy. Unlike VTON's body-centric masks, our method combines leg pose information with bounding box expansion from shoe detection to create adaptive masks. This allows seamless cross-shoe category try-on capabilities, for example, a model wearing long boots or flip-flops can be transformed into sneakers while maintaining anatomical alignment and realistic occlusion. As illustrated in the first row in Fig.~\ref{supp: fig: comparison}, this approach lays the foundation for robust, universal shoe try-on systems.

However, our mask generation strategy is not without its limitations. A failure case is presented in the second row of Fig.~\ref{fig: main}(b). Here, the region of the rear shoe is both heavily occluded and minimally visible, which leads to the failure of our keypoint and shoe detection modules. Consequently, this rear shoe area is erroneously classified as background, and only the front foot is designated as the inpainting region. This incomplete mask ultimately results in an unsuccessful virtual try-on. We contend that this is an inherent challenge for all methods reliant on an explicit masking strategy, and developing more robust solutions to handle such severe occlusion cases remains an important direction for our future work.

\paragraph{Shoe Expansion Phenomenon.} 
% \label{fig: motivation}
The volumetric expansion of footwear observed during the try-on process is a naturally occurring phenomenon. In real-world datasets, it is common to find that shoes with stiffer tongues tend to bulge outwards when worn, resulting in a slight expansion of the shoe's overall volume. This effect is illustrated in Fig.~\ref{fig: main}, where the third column clearly shows a prominent bulge in the tongue area, leading to body expansion. This expansion occurs stochastically in real-world data pairs, as exemplified in the third row of Fig.~\ref{fig: motivation}(a), where the shoe upper in the second column exhibits a slightly thicker contour. Consequently, by effectively fitting to the training data distribution, our model learns to replicate this real-world phenomenon, accurately rendering the corresponding deformations inherent to the try-on process.

\begin{table*}[h]
\centering
\caption{Shoe Category Statistics. We report the sample count for each subcategory, with its proportion indicated in parentheses.}
\begin{tabular}{|l|p{8cm}|}
\hline
\textbf{Primary Categories} & \textbf{Subcategory} \\
\hline
\textit{Casual Shoes}: 3158 (43.23\%) & Lifestyle Casual Shoes: 803 (10.99\%) \tabularnewline 
& Sneakers: 1428 (19.55\%) \tabularnewline
& Canvas Shoes: 141 (1.93\%) \tabularnewline
& Children Casual Shoes: 99 (1.36\%) \tabularnewline
& Men Casual Shoes: 687 (9.40\%) \\
\hline
\textit{Athletic Shoes}: 558 (7.64\%) & Running Shoes: 322 (4.41\%) \tabularnewline
& Basketball Shoes: 121 (1.66\%) \tabularnewline
& Training Shoes: 45 (0.62\%) \tabularnewline
& Football Shoes: 70 (0.96\%) \\
\hline
\textit{High-Heel Shoes}: 895 (12.25\%)  & Classical High-heel Shoes: 340 (4.65\%) \tabularnewline
& Ladies Casual Shoes: 401 (5.49\%) \tabularnewline
& Mary Jane Shoes: 154 (2.11\%) \\
\hline
\textit{Boots}: 1045 (14.31\%)  & Ankle Boots: 416 (5.69\%) \tabularnewline
& Chelsea Boots: 56 (0.77\%) \tabularnewline
& High Boots: 153 (2.09\%) \tabularnewline
& Snow Boots: 183 (2.51\%) \tabularnewline
& Martin Boots: 237 (3.24\%) \\
\hline
\textit{Sandals}: 279 (3.82\%)  & Flip Flops: 183 (2.51\%) \tabularnewline
& Beach Sandals: 36 (0.49\%) \tabularnewline
& Strap Sandals: 60 (0.82\%) \\
\hline
\textit{Dress Shoes}: 389 (5.33\%)  & Loafers: 209 (2.86\%) \tabularnewline
& Formal Leather Shoes: 180 (2.46\%) \\
\hline
\textit{Slippers}: 981 (13.43\%)  & Clogs: 82 (1.12\%) \tabularnewline
& Thong Slippers: 18 (0.25\%) \tabularnewline
& House Slippers: 881 (12.06\%) \\
\hline
\end{tabular}
\label{supp: tab: cate}
\end{table*}

\begin{table}
%%\vspace{-7mm}
  \centering
  \caption{Human scene statistics.}
  \label{supp: tab: scene}
 
  \begin{tabular}{lcc}
  \toprule
 Scene Types         & Number of samples        & Percentage     \\
\midrule
\textit{Top-down Foot}            & 1719        & 23.53\%                \\
\textit{Horizontal Foot}          & 4330        & 59.27\%                \\
\textit{Half-body Model}          & 833         & 11.40\%               \\
\textit{Full-body Model}          & 423         & 5.79\%               \\
\bottomrule
  \end{tabular}

\end{table}

%--------------------------------------------------------------------------

\section{Method Supplement}
\label{supp: sec: supplement_method}

\subsection{Preliminary}
\label{supp: sec: preliminary}

\paragraph{FLUX.1}  
Our ShoeFit is an extension of Stable Diffusion 3~\cite {rombach2022ldm} and FLUX.1~\cite{flux}, which are the most commonly used text-to-image diffusion models based on Flow Matching\cite{lipman2022flow} and DiT\cite{dit}. FLUX.1~\cite{flux} employs a variational autoencoder~\cite{kingma2013vae} (VAE) that consists of an encoder $\mathcal{E}$ and a decoder $\mathcal{D}$ to enable image representations in the latent space. It is also equipped with rotary positional embeddings (RoPE)\cite{su2024roformer} and denoise-text-stream attention layers to improvement performance. FLUX.1 implements an actual two-dimensional RoPE scheme for encoding spatial positions in the latent space:
\begin{equation}
\omega_d = \frac{1}{\theta^{2d/D}}, \text{for} D=0, 1, ..., D/2-1,
\end{equation}
where $\theta$ is typically set to 10000. The position encoding applies a rotation matrix:
\begin{equation}
\begin{bmatrix} 
\cos(\omega_d \cdot \mathbf{pos}) & -\sin(\omega_d \cdot \mathbf{pos}) \\
\sin(\omega_d \cdot \mathbf{pos}) & \cos(\omega_d \cdot \mathbf{pos})
\end{bmatrix}
\end{equation}
This rotation is applied to query and key vectors in the attention mechanism, enabling the model to capture relative positional relationships in the latent space.

\paragraph{Flow Matching}  
Flow matching~\cite{lipman2022flow} aligns the flow of information between noise $\epsilon$ and data distributions by optimizing a velocity field $u_t$, which progressively converts noise into data over time. This technique ensures that the generative model maps the noise distribution to the actual data distribution in a structured manner. The text encoders~\cite{radford2021clip} $\tau_{\theta}$ are employed to deal with the given text prompt $y$.  The flow matching loss is defined as follows:
\begin{equation}
\mathcal{L} = E_{t, p_t(z|\epsilon), p(\epsilon),y} \left[ \left\| v_\Theta(z, t, \tau_{\theta}(y)) - u_t(z|\epsilon) \right\|^2 \right].
\end{equation}
In this context, \( v_\Theta(z, t, \tau_{\theta}(y)) \) signifies the conditional velocity field determined by the weights of the neural network, while \( u_t(z|\epsilon) \) represents the vector field created by the model to delineate the probabilistic trajectory between the noise and actual data distributions. The symbol \( E \) stands for the expectation, which involves either integration or summation over time \( t \), latent variables \( z \), conditions \( y \), and noise \( \epsilon \). This expectation computes the mean of the squared differences for all conditions, ensuring that the model's performance is evaluated over numerous instances to yield a dependable estimate of its generative capability.

\subsection{LayeredRefAttention in Single-stream DiT Block}

\begin{wrapfigure}{r}{0.5\textwidth}

  \centering
  \includegraphics[width=0.48\textwidth]{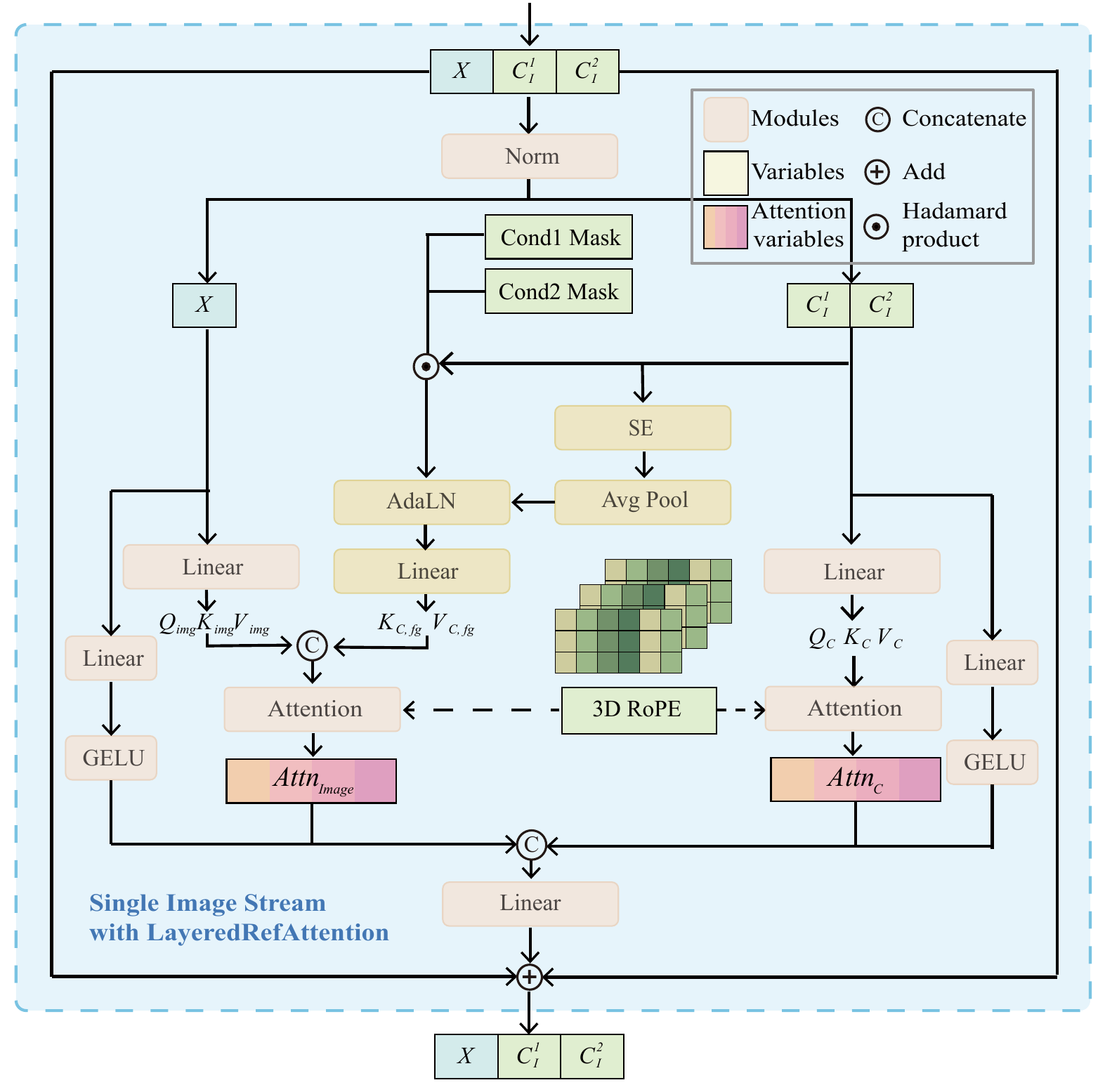}

  \caption{We provide how LayeredRefAttention is used within a single-stream DiT block.} 
  \label{supp: fig: single}

\end{wrapfigure}

The original FLUX is a text-to-image model composed of a series of stacked MM-DiT (double-stream) blocks, followed by a series of stacked single-stream DiT blocks. 
Due to space constraints, the framework of the LayeredRefAttention module shown in the main text is its structure within the double-stream DiT blocks. Here, we provide how it is used within a single-stream DiT block in Fig.~\ref{supp: fig: single}. Similar to the way in double-stream blocks, we employ a Squeeze-and-Excitation (SE) block ~\cite{hu2018squeeze} followed by global average pooling over spatial dimensions to compute channel-specific weights: 
\begin{equation}
P^{(n)}=AvgPool(SE(C_I^{(n)})\in \mathbb{R}^{b\times c}.
\end{equation}
Subsequently, we employ two linear layers $F(\cdot)$ to extract background modulation parameters and modulate foreground shoe features, filtering irrelevant environmental lighting and background reflection to ensure faithful material preservation of shoes as:
\begin{equation}
\beta_{scale}^{(n)}, \beta_{shift}^{(n)} = F(P^{(n)}), \quad C_{fg}^{(n)} = (1+\beta_{scale}^{(n)})\cdot (LN(C_I^{(n)} \odot M_{fg}^{(n)})) +  \beta_{shift}^{(n)},
\end{equation}
where $LN(\cdot)$ means layer normalization and $M_{fg}^{(n)}$ represents the binary shoe masks for conditioning product image. The subsequent operations are then performed as in regular single-stream DiT blocks.

\begin{figure}[t]
  \centering
\includegraphics[width=1.0\textwidth]{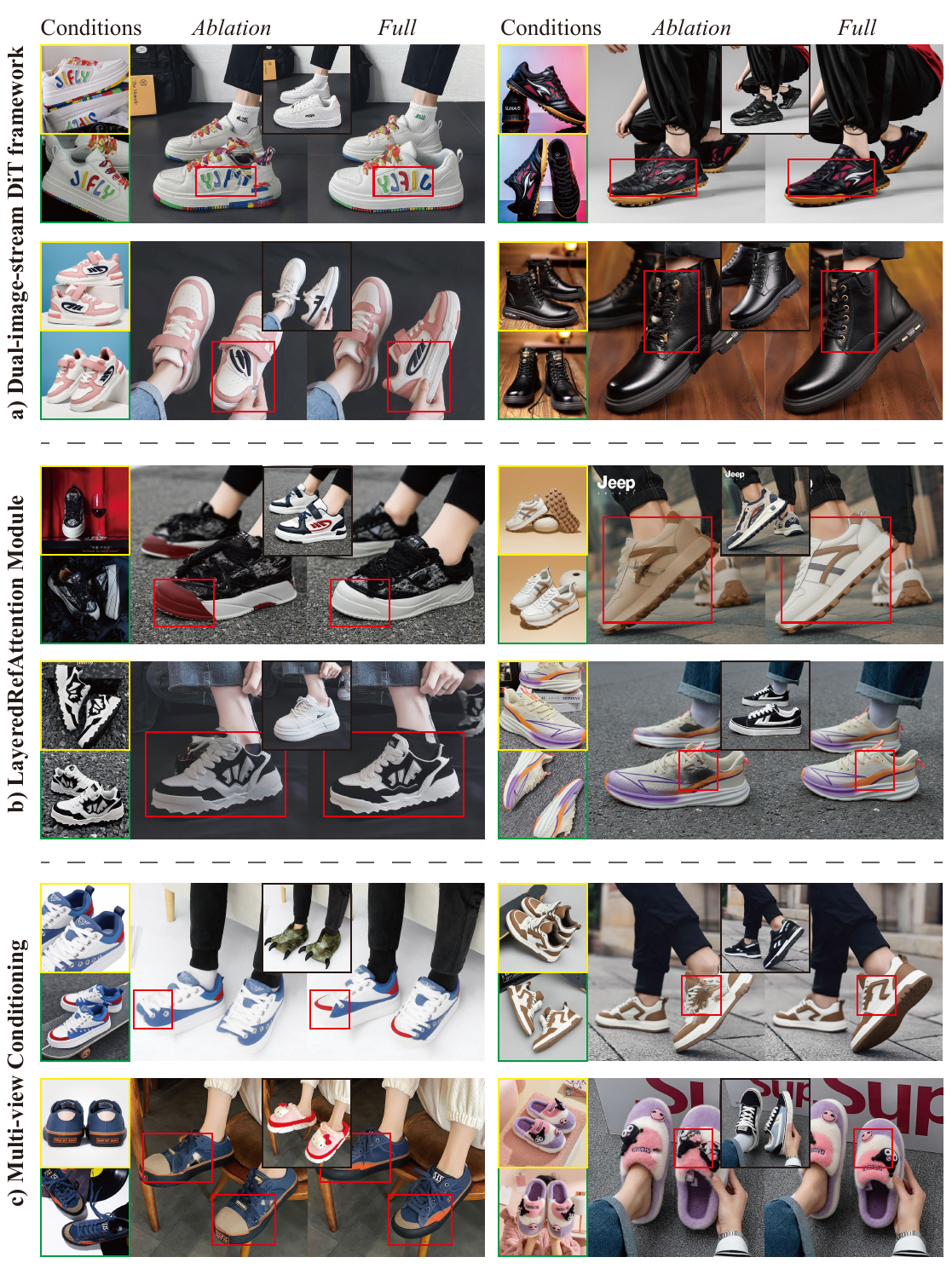}
  \caption{More visual results of the ablation study. We highlight the improvements by all three components of the method in red. Best viewed when zoomed in.} 
  \label{supp: fig: ablation}
\end{figure}

%--------------------------------------------------------------------------

\section{Implementation Supplement}
\label{supp: sec: imple_details}

\subsection{Baseline Training Details}
We here provide detailed descriptions of the training processes for the baseline models, focusing specifically on the training configurations. All models are trained on MVShoes at a resolution of $768\times 768$, utilizing 6305 pairs for training and 1000 pairs for testing.

For Flux.1 Fill~\cite{flux-fill}, we implement the training and inference pipeline by concatenating footwear images with human model images. In the trainable components, we apply LoRA~\cite{hu2022lora} with a rank of 64 to all attention modules in the model and a Flux ControlNet~\cite{zhang2023controlnet} comprising 6 single layers and 6 double layers, injecting the poses at every denoising step. The model is trained for 6 days on 8 80GB-A100 GPUs using DeepSpeed ZeRO-2, with a batch size of 4. We utilize the AdamW optimizer, setting a constant learning rate of 3e-5 for training, and operate the model on a single A100 GPU for 25 steps during inference.

For Flux.1 Redux~\cite{flux-redux}, we adopt the same training settings as those used for FLUX.1-Fill. Additionally, we incorporate two linear layers for projecting SigLip~\cite{zhai2023sigmoidlosslanguageimage} features of the product shoe images, training these two layers concurrently with the attention LoRAs and ControlNet. The parameters of SigLip remain frozen throughout the training process. We retain the same inference settings as employed in FLUX.1-Fill.

For OOTDiffusion~\cite{xu2024ootdiffusion}, we execute the training and inference pipelines based on the official code. We also implement ControlNet to facilitate pose injection. The model is trained for 6 days on 4 80GB-A100 GPUs with DeepSpeed ZeRO-2, at a batch size of 8. During inference, the model is run on a single A10 GPU for 25 steps.

For CatVTON-Flux-Lite-2V, we maintain the text stream for FLUX-lite and concatenate footwear images with human model images. The model undergoes training for 7 days on 12 80GB-A100 GPUs utilizing DeepSpeed ZeRO-2, with a batch size of 2. All other training and inference configurations align with those of ShoeFit-2V.

\subsection{Data Augmentation} 
We have implemented data augmentation techniques that could potentially enhance the model's generalization ability as well as its fidelity performance. Specifically, the data augmentation operations include (a) horizontal flipping of images, (b) resizing footwear and human figures through padding (up to 10\% of the image size), (c) randomly adjusting the image's hue within a range of -5 to +5, and (d) randomly adjusting the image's contrast within a specified range (between 0.8 and 1.2 times the original contrast). Each of these operations occurs independently with a 50\% probability. Moreover, these operations are simultaneously applied to both the footwear and model images.

\subsection{LayeredRefAttention Hyperparameters}
In the LayeredRefAttention layer, we primarily introduced a new Linear layer and an SE block. The Linear layer for the foreground follows the same dimensions as other linear layers within the module, which is the \textit{hidden size}. The SE block takes input features with the same dimensions of \textit{hidden size}, and internally, we use a compression rate of $reduction=4$ for SE Block, as $\text{Linear}(channels, channels / 4)$, $\text{ReLU}()$, $\text{Linear}(channels / 4, channels)$.

%--------------------------------------------------------------------------

\section{More visual results}
\label{supp: sec: more_results}

Fig.~\ref{supp: fig: ablation} provides more results about the ablation study. We highlight the improvements by all three components of the method in red. 

\begin{figure}[t]
  \centering
\includegraphics[width=1.0\textwidth]{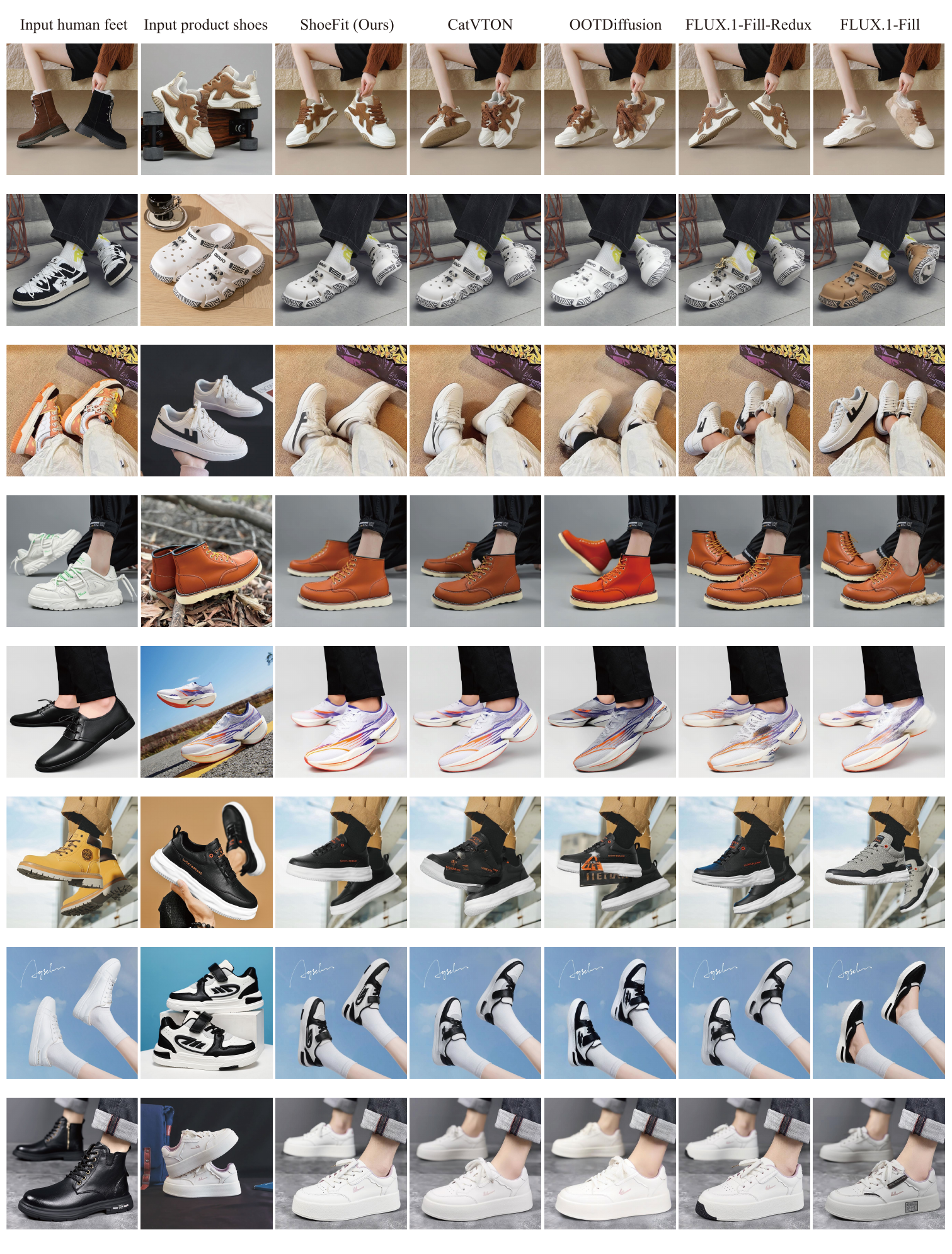}
  \caption{More visual comparisons on the MVShoes by ShoeFit. Best viewed when zoomed in.} 
  \label{supp: fig: comparison}
\end{figure}

\noindent Fig.~\ref{supp: fig: comparison} provides more results on MVShoes for comparisons between the baselines and ShoeFit. For a fair
comparison, we report the results of the single-view conditioning version of all methods. Our method substantially improves rendering fidelity and robustness under challenging real-world product shoes, establishing a new benchmark in high-fidelity footwear try-on synthesis. 

\begin{figure}[t]
  \centering
\includegraphics[width=1.0\textwidth]{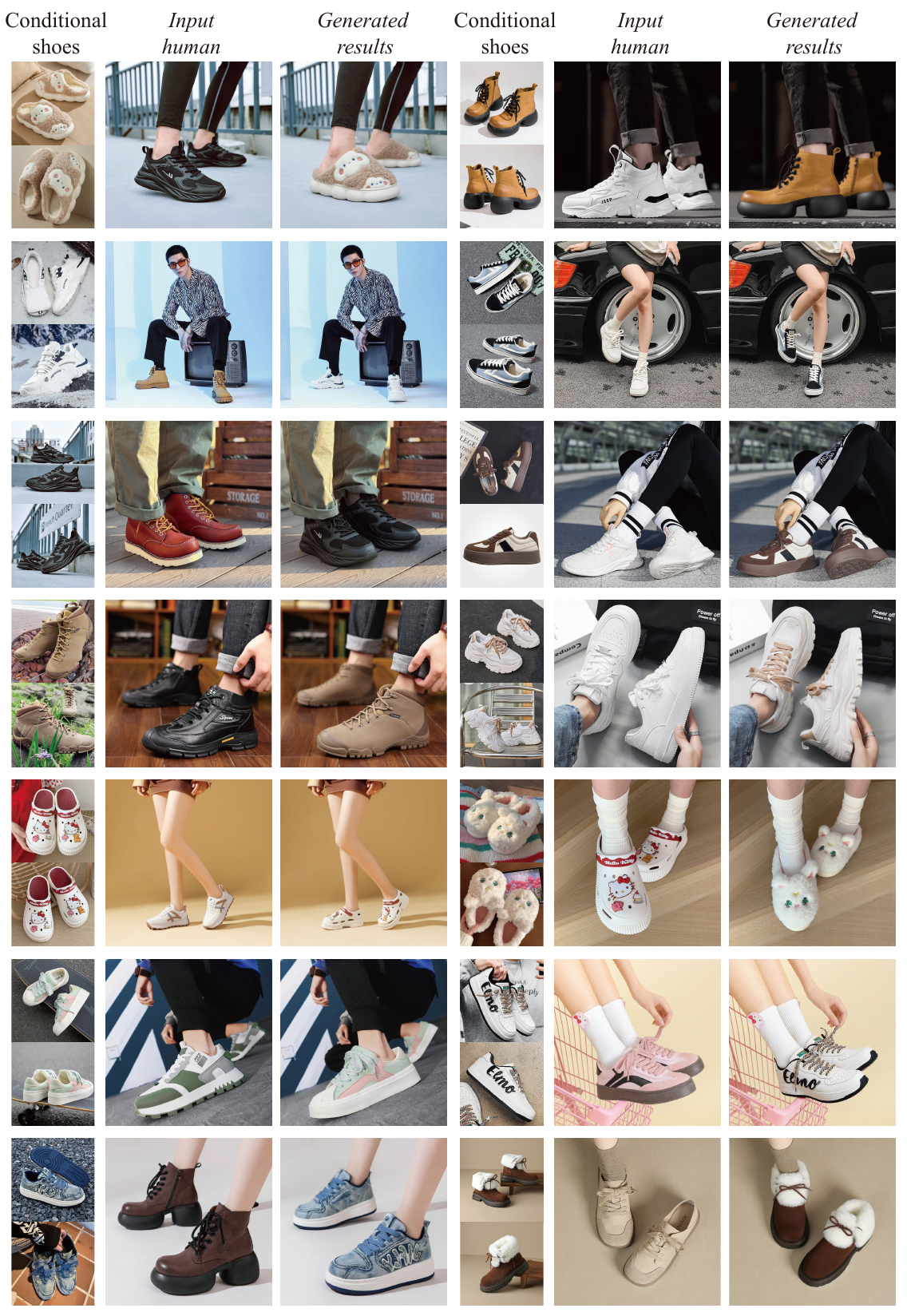}
  \caption{More visual results on the MVShoes by ShoeFit. Best viewed when zoomed in.} 
  \label{supp: fig: showcase}
\end{figure}

\noindent Fig.~\ref{supp: fig: showcase} provides more results on MVShoes for inspection to demonstrate that ShoeFit synthesizes high-fidelity and detail-faithful try-on results. 

\begin{figure}[t]
  \centering
\includegraphics[width=1.0\textwidth]{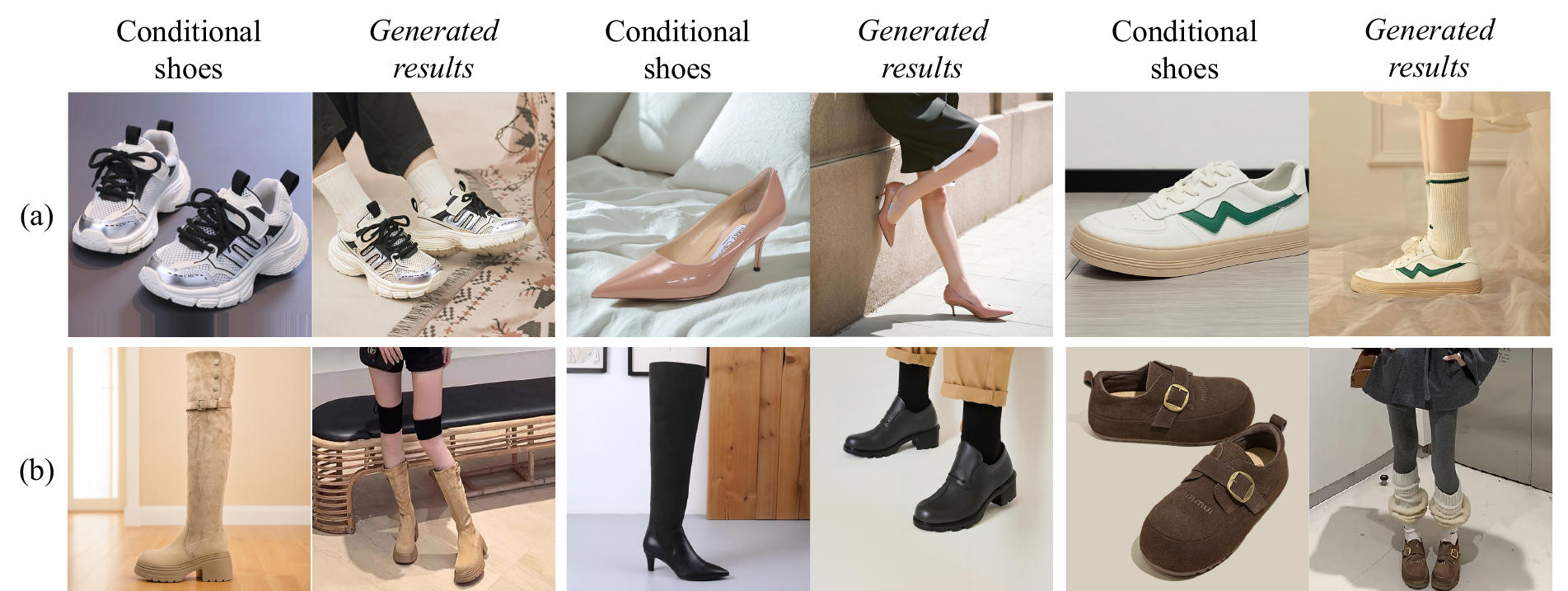}
  \caption{More visual results on two primary failure modes of our method: (a) structural failure observed during early training stages, and (b) artifacts arising from inaccurate masks in long footwear scenarios.} 
  \label{supp: fig: badcase}
\end{figure}

\noindent Fig.~\ref{supp: fig: badcase} details two key limitations of our model. First, in the early training phase, the model has not yet mastered spatial reasoning, resulting in structurally distorted outputs (row (a)). Second, generating precise masks for tall footwear remains a challenge. As shown in the row (b), inaccurate masks for items like boots can cause significant visual artifacts or fitting errors.

\begin{figure}[t]
  \centering
\includegraphics[width=1.0\textwidth]{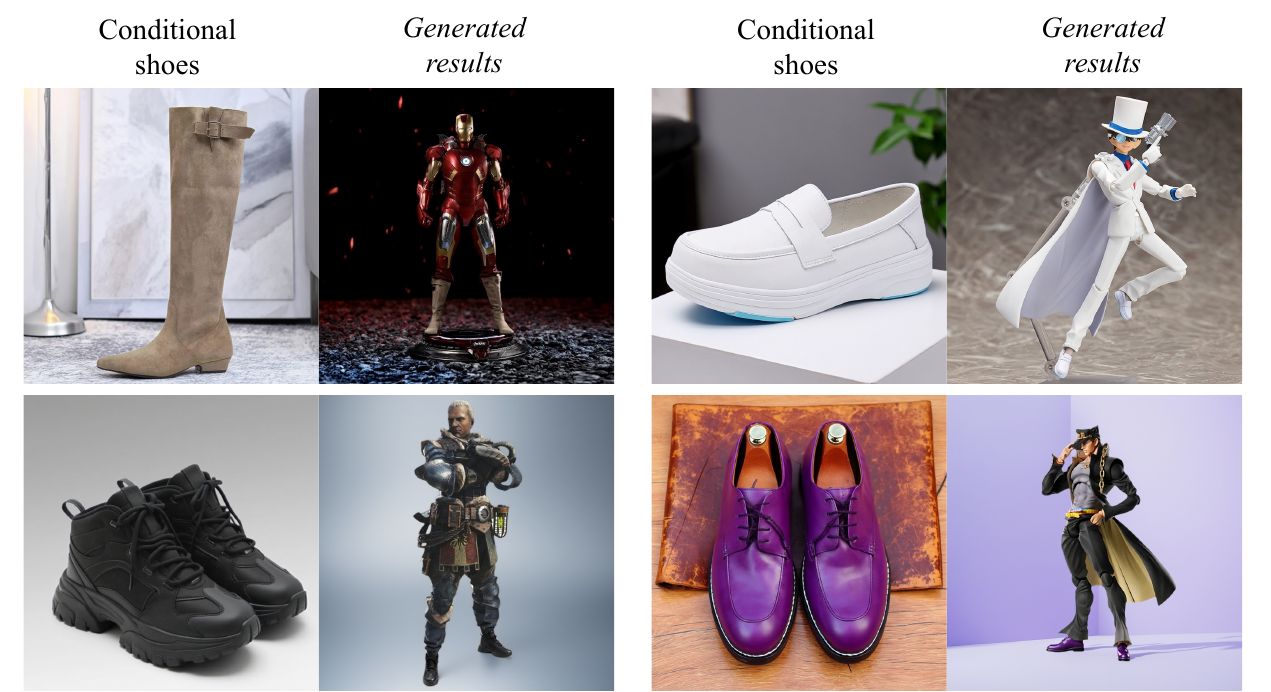}
  \caption{More visual results on out-of-distribution (OOD) samples not present in the MVShoes dataset. The high-quality try-on results demonstrate the model's notable stability and strong generalization capabilities.} 
  \label{supp: fig: acg}
\end{figure}

\noindent Fig.~\ref{supp: fig: acg} illustrate more results on out-of-distribution (OOD) samples not present in the MVShoes dataset. The high-quality try-on results demonstrate the model's notable stability and strong generalization capabilities.

\clearpage

%%%%%%%%%%%%%%%%%%%%%%%%%%%%%%%%%%%%%%%%%%%%%%%%%%%%%%%%%%%%

% \newpage
% \appendix

% \input{sec/X_supp}

%%%%%%%%%%%%%%%%%%%%%%%%%%%%%%%%%%%%%%%%%%%%%%%%%%%%%%%%%%%%

% \newpage
% \input{sec/X_checklist}

\end{document}